\documentclass[11pt]{article}

\usepackage[margin=1in]{geometry}
\usepackage{authblk}

\usepackage{custompkg}
\usepackage{customdef}
\usepackage[caption=false,font=footnotesize]{subfig}
\usepackage{nicefrac}
\usepackage{algorithmic}
\usepackage{xspace}
\usepackage[noadjust]{cite}

\usepackage{pgfplots}
\usepgfplotslibrary{external}

\usepackage{tikz}
\tikzsetexternalprefix{assets/figures/}
\tikzexternaldisable

\usepackage{hyperref}

\newcommand{\email}[1]{\href{mailto:#1}{#1}}

\usetikzlibrary{
  arrows,
  arrows.meta,
  backgrounds,
  decorations.text,
  fit,
  math,
  patterns,
  positioning,
  shapes,
  trees
}

\newcommand{\R}{\set{R}}

\renewcommand{\secref}[1]{Section~\ref{#1}}

\newcommand{\hogwild}{\textsc{Hogwild!}\xspace}
\newcommand{\sgd}{\textsc{sgd}\xspace}
\newcommand{\sag}{\textsc{sag}\xspace}
\newcommand{\saga}{\textsc{saga}\xspace}
\newcommand{\svrg}{\textsc{svrg}\xspace}
\newcommand{\sdca}{\textsc{sdca}\xspace}
\newcommand{\iag}{\textsc{iag}\xspace}
\newcommand{\pushsum}{\textsc{Push-Sum}\xspace}

\newcommand{\asaga}{\textsc{asaga}\xspace}
\newcommand{\miso}{\textsc{miso}\xspace}
\newcommand{\finito}{\textsc{finito}\xspace}
\newcommand{\sgp}{\textsc{sgp}\xspace}
\newcommand{\osgp}{\textsc{osgp}\xspace}
\newcommand{\agp}{\textsc{agp}\xspace}
\newcommand{\allreduce}{\textsc{AllReduce}\xspace}

\newcommand{\taubar}{\tau_{\max}}

\title{Advances in Asynchronous Parallel and Distributed Optimization}
\author{Mahmoud~Assran, Arda~Aytekin, Hamid~Feyzmahdavian, Mikael~Johansson, and Michael~Rabbat%
\thanks{Email: \email{mahmoud.assran@mail.mcgill.ca}, \email{arda@aytekin.biz}, \email{hamid.feyzmahdavian@se.abb.com}, \email{mikaelj@kth.se}, \email{mikerabbat@fb.com}}%
}

\begin{document}
\maketitle

\begin{abstract}
Motivated by large-scale optimization problems arising in the context of machine learning, there have been several advances in the study of asynchronous parallel and distributed optimization methods during the past decade. Asynchronous methods do not require all processors to maintain a consistent view of the optimization variables. Consequently, they generally can make more efficient use of computational resources than synchronous methods, and they are not sensitive to issues like stragglers (\ie, slow nodes) and unreliable communication links. Mathematical modeling of asynchronous methods involves proper accounting of information delays, which makes their analysis challenging. This article reviews recent developments in the design and analysis of asynchronous optimization methods, covering both centralized methods, where all processors update a master copy of the optimization variables, and decentralized methods, where each processor maintains a local copy of the variables. The analysis provides insights as to how the degree of asynchrony impacts convergence rates, especially in stochastic optimization methods.
\end{abstract}

\section{Introduction}
\label{sec:intro}

Since the slowing of Moore's scaling law, parallel and distributed computing have become a primary means to solve large computational problems. Much of the work on parallel and distributed optimization during the past decade has been motivated by machine learning applications. The goal of fitting a predictive model to a dataset is formulated as an optimization problem that involves finding the model parameters that provide the best predictive performance. During the same time, advances in machine learning have been enabled by the availability of ever larger datasets and the ability to use larger models, resulting in optimization problems potentially involving billions of free parameters and billions of data samples~\cite{radford2019language,liu2019roberta,mahajan2018exploring}.

There are two general scenarios where the use of parallel computing resources naturally arises. In one scenario, the data is available in one central location (\eg, a data center), and the aim is to use parallel computing to train a model faster than would be possible using serial methods. The ideal outcome is to find a parallel method that achieves \emph{linear scaling}, where the time to achieve a solution of a particular quality decreases proportionally to the number of processors used; \ie, doubling the number of parallel processors reduces the compute time by half. However, unlike serial methods, parallel optimization methods generally require coordination or communication among multiple processors.

In the second scenario, which is receiving increasing interest, the data is widely distributed (\eg, residing on users' devices), and the goal is to train a model using all of the data without collecting it in one location. The motivation to process the data in a distributed way may be for privacy reasons, and it may also be too expensive (in terms of communication time and bandwidth) to communicate the data. 
Although this survey primarily focuses on the first scenario, many of the results and methods discussed can be readily applied in the second.

Parallel and distributed algorithms may be classified as synchronous or asynchronous. Synchronous algorithms require that the processors \emph{serialize} after every update, so that every processor always has a consistent view of optimization variables. This generally makes serial algorithms easier to analyze, implement, and debug, and consequently synchronous algorithms have been more widely studied. However, synchronization may also lead to poor utilization of computational resources. If one processor is slower than the others, they must all wait \emph{idling} at the serialization point until the slowest processor catches up, resulting in wasted compute cycles. 

Asynchronous algorithms do not impose a global synchronization point at the end of each iteration. Consequently, each processor may have a different view of the optimization variable when performing local computations. Properly accounting for these inconsistencies complicates the analysis of asynchronous iterative algorithms. Also, because their execution is non-deterministic, asynchronous algorithms can be challenging to implement and debug. The appeal of asynchronous methods is that they indeed can make more efficient use of computational resources, resulting in faster wall-clock convergence.

\subsection{Historical context}

The dynamics of asynchronous iterations are much richer than their synchronous counterparts, and quantifying the impact of asynchrony on the convergence times of iterative algorithms is challenging.  Some of the first results on the convergence of asynchronous iterations were derived by Chazan and Miranker~\cite{Chazan:69} who were motivated by encouraging empirical results for parallel and asynchronous linear equation solvers~\cite{Ros:67}. Their theoretical results considered linear iterations under bounded asynchrony. Several authors have extended this work to nonlinear iterations involving maximum norm contractions (\eg,~\cite{Baudet:78}) and for monotone iterations (\eg,~\cite{Bertsekas:87}). Powerful convergence results for broad classes of asynchronous algorithms, including maximum norm contractions and monotone mappings, under different assumptions on communication
delays and update rates were presented by Bertsekas~\cite{Bertsekas:83}
and in the celebrated book of Bertsekas and Tsitsiklis~\cite{Bertsekas:89}. An important insight in this line of work is that asynchrony can be modelled as time-varying update rates and information delays with respect to a global ordering of events in the system.

The framework of~\cite{Bertsekas:89} defines two models of asynchrony: \emph{totally asynchronous} and \emph{partially asynchronous algorithms}. In totally asynchronous algorithms the information delays may grow arbitrarily large, and therefore the best one can expect is asymptotic convergence. In partially asynchronous algorithms, both inter-update times and information delays remain bounded and algorithms may converge to a target accuracy in finite time.

Although the framework for modeling asynchronous algorithms in~\cite{Bertsekas:89} is both powerful and elegant, the most concrete results consider totally asynchronous iterations. In particular, pseudo contractions in the block-maximum norm are shown to converge when executed in a totally asynchronous manner; however, in machine-learning applications, first-order methods rarely result in maximum-norm contractions. For example, for convex quadratic optimization problems, the gradient descent iterations are maximum norm contractions only if the Hessian is diagonally dominant.
The asymptotic nature of these results also means that they do not characterize how the amount of asynchrony impacts convergence times.

In contrast, we will show below that many important optimization algorithms for machine-learning tolerate some level of asynchrony and that it is possible to quantify how asynchrony affects the number of iterations required to find a solution with a given target accuracy. Such results provide important insight into engineering trade-offs between the longer iteration times in synchronous systems and additional (but faster) iterations in asynchronous implementations. A challenge comes from the fact that many optimization methods used for deep learning are inherently stochastic; \eg, they use sampling to approximate the gradient when determining the direction in which to update the model parameters. The results available in~\cite{Bertsekas:89} for partially asynchronous methods are difficult to apply in the stochastic setting, where randomized update orders and workloads affect update frequencies and computation times.
To address these challenges, a new wave of research on analysis and design of asynchronous and distributed optimization algorithms has emerged, and significant theoretical and technological advances have recently been made.

\subsection{This article}

This article surveys advances in the field of asynchronous and parallel distributed optimization made in the past decade. Since the work we survey is mainly motivated by applications from machine learning, we review background material on this topic in \secref{sec:background}. Then, in \secref{sec:architectures}, we introduce necessary background on parallel and distributed computing systems.

The parallel and distributed optimization methods we survey can be divided into two categories. \emph{Centralized} methods, which we discuss in \secref{sec:shared}, maintain one master copy of the optimization variables. In one iteration of these methods, a processor reads the master variables, performs a local computation, and then updates the master copy. However, between the time a processor reads the master variables and then updates them, other processors may have already performed updates. In extreme cases, the master variables may be updated simultaneously by multiple processors while they are being read by others. 

\emph{Decentralized} methods, discussed in \secref{sec:multi-agent}, form the second category. In decentralized methods there is no master copy; rather, each processor maintains a local copy of the optimization variables. Processors update their local copies and synchronize them with other processors directly. These methods are typically implemented on distributed-memory systems, although they may also be applicable in large shared-memory systems with non-uniform memory access, where the time it takes a given processor to access different locations in memory depends on the physical distance between the memory location and the processor.

Throughout this article we focus on first-order methods, which only make use of gradient information, since they are the most widely-used methods for training machine learning models today~\cite{bottou2007tradeoffs,bottou2018optimization}. We also focus on stochastic optimization methods, where gradient information is obtained by sampling a subset of data points. Newton-type methods, which make use of second-order derivatives and curvature, have not received wide adoption because they involve computing and storing the Hessian matrix or an approximation thereof, which requires significant computation and memory when the problem dimension is large, as is typical of many machine learning problems. Also, Newton-type methods tend to be more sensitive to noise and stochasticity.

When discussing both centralized and decentralized methods, we survey existing algorithms and their convergence guarantees, and we also discuss the main analysis techniques. We emphasize results for the partially asynchronous model, where information delays are bounded. In \secref{sec:multi-agent} we include a numerical example illustrating how asynchronous decentralized algorithms may be used for training deep neural-networks. We conclude in \secref{sec:conc} with a discussion of open problems and directions for future work.

\subsection{Other applications}

Throughout the rest of this article, we will use example applications from machine learning to illustrate asynchronous optimization methods. We note in passing that asynchronous parallel and distributed optimization methods are also relevant to a variety of other applications, including the operation of distributed infrastructure systems such as power systems~\cite{ramanan2019asynchronous} and water distribution, the internet-of-things (IoT), smart grids, networks of autonomous
vehicles, and wireless communication networks~\cite{scutari2008asynchronous}.

\section{Background}
\label{sec:background}

\subsection{Optimization}
This article focuses on optimization methods to solve the problem
\begin{align}
\label{eq:optimization-problem}
  \begin{aligned}
    & \minimize_{w \in \mathbb{R}^{d}} & & f(w) \,.
  \end{aligned}
\end{align}
A point $w \in \R^d$ is called a local minimizer of the objective function $f$ if $f(w) \le f(w')$ for all other points $w'$ in a neighborhood of $w$, and $w^\star$ is called a global minimizer if $f(w^\star) \le f(w')$ for all $w' \in \R^d$.

Serial iterative optimization methods start from an initial point $w^{(0)}$ and generate a sequence $w^{(1)}, w^{(2)}, \dots$ by performing update steps. The performance of an optimization method is measured in terms of the number of steps required to reach a solution of a certain quality. Performance guarantees depend on the particular method, the assumed characteristics of the objective function $f$, and how it can be evaluated.

Update steps, going from $w^{(k)}$ to $w^{(k+1)}$, typically make use of the (negative) gradient $- \nabla f(w^{(k)})$ at $w^{(k)}$, which points in the direction that decreases the objective function $f$ in a small neighborhood around $w^{(k)}$. The objective function is said to be \emph{smooth} if the gradient changes gradually; formally, there is a constant $L > 0$ such that
\begin{align*}
\norm{\nabla f(w) - \nabla f(w')} \le L \norm{w - w'}
\end{align*}
for all points $w, w' \in \R^d$. The smaller $L$, the smoother the function $f$, and consequently first-order optimization methods can take larger steps while still ensuring convergence. A point $w \in \R^d$ is called a \emph{stationary point} if $\nabla f(w) = 0$; in this case the gradient does not point in any direction, so $w$ may be a (local or global) minimizer. Non-smooth problems are more challenging because even a small change in parameter space may result in a drastic change in the gradient direction.

An important class of optimization problems is related to the notion of convexity. Formally, $f$ is convex if for any $\alpha \in (0,1)$ it holds that
\begin{align*}
f(\alpha w + (1 - \alpha)w') \le \alpha f(w) + (1 - \alpha) f(w').    
\end{align*}
When $f$ is convex, every stationary point of $f$ is a global minimizer of $f$, and so iterative methods that converge to stationary points are guaranteed to find a global minimizer. In general, there may be many points that minimize $f$. When $f(w) - \frac{\mu}{2} \norm{w}_2^2$ is also convex, for some parameter $\mu > 0$, we say that $f$ is \emph{strongly convex}, and we refer to $\mu$ as the strong convexity parameter. Strongly convex functions have a unique global minimizer that is also the unique stationary point of $f$, and they are also easier to optimize than functions which are only convex. We refer the interested reader to \cite{nesterov2004introductory} for details.

\subsection{Machine learning}

Machine learning methods train a parameterized model to make predictions on a set of data with the goal that the model makes accurate predictions on never-before seen data. For example, in an image classification task, the model is shown an image and asked to predict which class, among a finite but possibly large set of classes, best describes the image content (\eg, dog, cat, human, \dots). Similarly, in a document classification task, the model is given a text and asked to predict which class best describes its content (\eg, in which section of the newspaper the text appeared).

Training of such models is typically formulated as an optimization problem on the form~\eqref{eq:optimization-problem} with
\begin{equation} \label{eq:erm}
f(w) = \frac{1}{m} \sum_{i=1}^m f_i(w) + r(w)
\end{equation}
and $f_i(w) = \ell(p(x_i; w), y_i)$. Here, $\{(x_i, y_i)\}_{i=1}^m$ denotes a collection of $m$ training samples, each consisting of an input $x_i$ and target $y_i$. The goal is to learn the parameters $w$ of a model $p(x; w)$ so that $p(x_i; w)$ matches $y_i$ well on the training data. The loss function $\ell$ measures how well a prediction $p(x_i; w)$ matches the target $y_i$. 
When $d > m$ or if the model class $\{p(x;w) : w \in \R^d\}$ is otherwise rich/expressive, one may use a regularization function $r(w)$ to avoid over-fitting the training data. The regularization function can also be used to impose other constraints on the model parameters $w$. 

This framework can describe a variety of common machine learning settings. For example, for a binary classification problem where $y_i \in \{0, 1\}$, using the model $p(y_i=1 | x_i; w) = \big(1 + \exp(- \dotprod{w}{x_i})\big)^{-1}$, loss $\ell(p, y_i) = -y_i \log(p) - (1 - y_i)\log(1 - p)$, and regularizer $r(w) = \frac{1}{2}\norm{w}_2^2$ corresponds to the popular $\ell_2$-regularized logistic regression method~\cite[Section~8.3]{murphy2012machine}; the resulting optimization problem is smooth and strongly convex. Other methods, including support vector machines, least-squares regression, and sparsity-inducing $\ell_1$-regularized methods can all be cast in this framework as convex optimization problems~\cite{shalevShwartz2014understanding}.

\emph{Deep neural networks} (DNNs) currently achieve state-of-the-art performance for the majority of machine learning tasks. The details of the DNN models $p(x; w)$ and loss functions $\ell$ are beyond the scope of this paper; the interested reader is referred to~\cite{goodfellow2016deep}. From the view of asynchronous distributed optimization, training DNNs also fits into the framework of~\eqref{eq:erm}, but the resulting optimization problem is typically not convex.

\subsection{Optimization methods for large scale learning}

An optimization problem with objective~\eqref{eq:erm} can become ``large-scale'' in a few different ways: the number of training samples $m$ can be large, the dimension of the training inputs $x$ and/or targets $y$ can become large, and the dimension $d$ of the optimization variable can be large. To handle the large $m$ setting, it is common to sample a smaller subset of samples at each update. In some cases, to handle the large $d$ setting one may also use coordinate descent methods that only update a subset of the coordinates at each update.

The iterative methods we consider all have the following general form: given an initial point $w^{(0)}$, repeat for $k \ge 0$,
\begin{equation} \label{eqn:first_order_update}
	w^{(k+1)} = w^{(k)} - \gamma^{(k)} s^{(k)},
\end{equation}
where $\gamma^{(k)} \in \R_+$ is a step-size, and $s^{(k)} \in \R^d$ is a search (or update) direction. The full gradient\footnote{To simplify the discussion throughout, we assume the loss and regularization functions are continuously differentiable, and will talk about gradient methods. We make specific remarks about modifications to handle non-smooth functions below.} of the objective $f$ in \eqref{eq:erm} may be expensive to compute (\eg, if $m$ or $d$ is large), so we focus on algorithms that approximate the gradient $\nabla f(w^{(k)})$ by a quantity which is easier to compute and which can be efficiently computed in parallel.

\subsection{Stochastic gradient descent}
\label{sec:synchronous-sgd}

To simplify the discussion for now, suppose that there is no regularizer; \ie, $r(w) = 0$. The classical \emph{stochastic gradient descent} (\sgd) method~\cite{robbins1951stochastic,bottou2018optimization} considers random search directions $s^{(k)}$ which are equal to the gradient in expectation and have bounded second moment, i.e.,
$\mathbb{E}[s^{(k)} \mid w^{(k)}] = \nabla f(w^{(k)})$ and $\mathbb{E}[ \|s^{(k)} - \nabla f(w^{(k)})\|^2 \mid w^{(k)}] \le \sigma^2$, where $\sigma^2 < \infty$ is assumed to be given.

For optimization problems of the form~\eqref{eq:erm}, one natural way to obtain random search directions is to use
\begin{align} \label{eqn:sgd-search-direction}
s^{(k)}=\nabla f_{i^{(k)}}(w^{(k)}),    
\end{align}
where $i^{(k)}$ is drawn uniformly at random from $\{1, \dots, m\}$. By only evaluating the gradient of a single component function, the computational cost per iteration is effectively reduced by a factor $m$.

The \sgd method~\eqref{eqn:first_order_update}--\eqref{eqn:sgd-search-direction} is inherently serial: the gradient computations take place on a single processor which needs access to the whole dataset. The desire to have faster methods for training larger models on larger datasets has resulted in a strong interest
in developing parallel optimization algorithms that are able to split the data and distribute the computation across multiple processors or multiple servers. 

A common practical solution for parallelizing stochastic gradient methods and reducing the stochastic variance is to employ mini-batches. Mini-batch \sgd method evaluates a subset $I^{(k)} \subseteq \{1, \dots, m\}$ of $b$ component gradients in parallel and uses the search direction
\begin{align} \label{eqn:mini-batch-direction}
s^{(k)} &= \frac{1}{b}\sum_{i \in I^{(k)}} \nabla f_i(w^{(k)}).
\end{align}
If we have $n \le b$  processors working in parallel, the hope is that we can compute $s^{(k)}$ roughly $n$ times faster than a single processor. In addition, by averaging $b$ component gradients, we reduce the variance of the search direction by a factor of $1/b$ per iteration. These factors combine to allow parallel mini-batch \sgd algorithms to achieve near-linear speedup in the number of compute nodes~\cite{Dekel:12}.

However, the parallel \sgd method as described above is \emph{synchronous}: the compute nodes all read the current decision vector $w^{(k)}$, evaluate the gradient of their assigned component function(s) at $w^{(k)}$, and then update the current iterate. Once all $b$ gradient updates have been applied to the decision vector, the algorithm can proceed to the next iteration.

Below we will discuss a variety of ways in which methods of the form~\eqref{eqn:first_order_update} can be made to run asynchronously. First, we review key aspects of parallel and distributed computing architectures, that will inform the subsequent discussion.

\section{Architectures}\label{sec:architectures}

Advances in computing hardware and communication infrastructures along with the
emergence of virtualization and container technologies have enabled a multitude
of options for affordable large-scale computing. High-performance
computing (HPC) environments traditionally available only at supercomputing
centers are now easily accessible as commoditized cloud services provided by
many companies. Similarly, hardware accelerators such as tensor processing units
(TPUs), general-purpose graphics processing units (GPUs) and multi-core central
processing units (CPUs), together with generous access to high-bandwidth memory
and storage have increased the compute capabilities of traditional servers by
many orders of magnitude. Finally, enabling technologies such as the 5G make it
possible to deploy and interconnect low-powered compute nodes to jointly collect
and process data on the \emph{edge}.

Obtaining the best possible performance from such compute resources relies on our
ability to parallelize the computations across multiple computational units (cores, devices, clouds).
Different compute resources can work concurrently using their local copies of
the model parameters and local datasets to speed up the computation of
(stochastic) gradients. However, simultaneously updating the shared model
parameters in \eqref{eqn:first_order_update} using these local gradients results
in undefined behaviour, and thus, has to be \emph{serialized}. There are two
different approaches to serializing simultaneous updates. \emph{Synchronous}
approaches mandate that \emph{all} compute nodes arrive at a \emph{barrier} to
exchange their local updates before proceeding. This makes it possible for all
the nodes to have the same view over the shared parameters at all times as if
the algorithm is running \emph{serially}. The main disadvantage of this approach
is that the overall performance of the algorithm depends on the \emph{slowest}
compute node. \emph{Asynchronous} approaches, on the other hand, let the compute
nodes update the shared parameters at their own pace. This makes it possible for
faster nodes to progress \emph{without} needing to wait for the slower nodes,
which results in better performance. However, the main challenge in this setting
is to incorporate the stale updates coming from slower nodes to the shared model
parameters.

\begin{figure}[t]
  \centering
  \includegraphics[width = \linewidth]{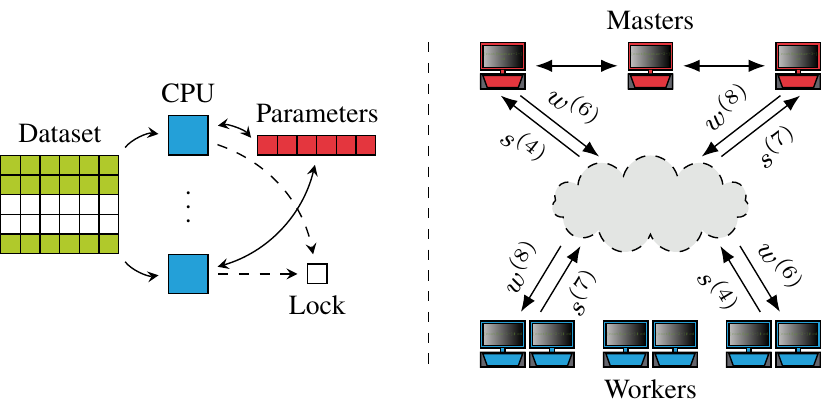}
  \caption{(Left) Shared-memory architecture where multiple CPUs read
    simultaneously from a dataset and update the parameters by obtaining a lock.
    (Right) Distributed-memory architecture where shared parameters are kept in
    centralized masters, and workers send their updates asynchronously.}
    \label{fig:architectures}
\end{figure}

We can categorize parallel computing architectures into two classes based on how
the simultaneous updates are serialized. \emph{Shared-memory architectures} span
compute resources such as many-core CPUs and hardware accelerators, which have
many compute nodes that share the same physical memory space
(Figure~\ref{fig:architectures}, left). Even though the nodes share the same
physical memory, non-uniform memory access (NUMA) designs and deep cache
hierarchies in today's architectures invalidate the assumption that nodes have
immediate access to a memory region (see, \eg, \cite{Chen2018,IMP:19,
Thakker2019,Blleloch2018,Kislal2016,Wszola2019} that revisit algorithms and take
this issue into account). In shared-memory architectures, all the serialization
primitives are in the same physical space. Synchronous operations are usually
implemented using \emph{semaphores}, where nodes signal their presence and
proceed with their next task only after every other node has also arrived.
Asynchronous operations, on the other hand, are usually implemented in one of
two ways: by mutual exclusion (\emph{mutex}) locks or \emph{atomic operations}.
Mutex locks are generally used to protect the shared model parameters \emph{as a
whole} during simultaneous accesses, which results in \emph{consistent} views
over the parameters. Atomic operations, on the other hand, protect the
\emph{individual} elements of the shared parameters, and thus, result in
\emph{inconsistent} views over the parameters as a whole. 

\begin{figure}
  \centering
  \includegraphics[width = 0.75\linewidth]{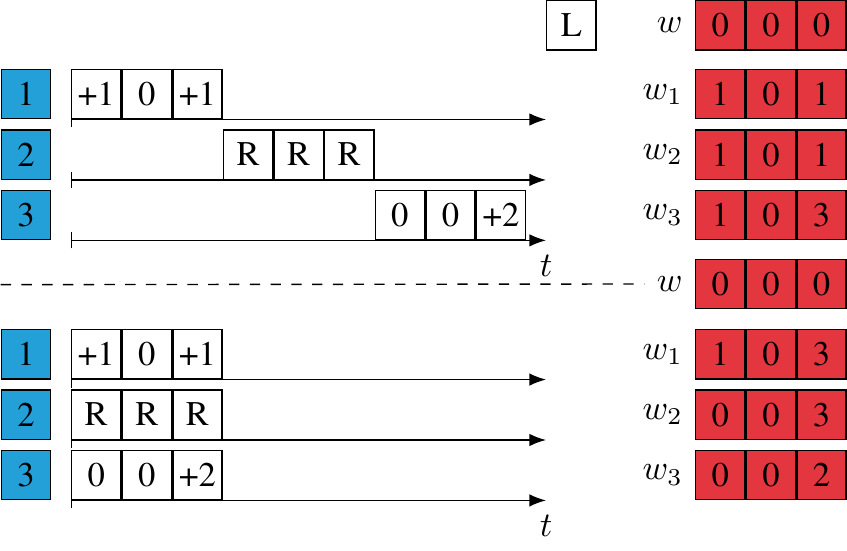}
  \caption{(Top) Serializing access to shared parameters using a mutex lock (L).
    (Bottom) Serializing access to individual elements using atomic operations.
    Different compute nodes' local updates are shown on the time axis for
    visualization purposes. Only the non-zero updates are applied, and thus,
    take time.}\label{fig:inconsistent}
\end{figure}

To better understand this, let us consider a scenario when two nodes are trying to update the shared
parameter vector $w = {[0,0,0]}^{\top}$ while another one is trying to read it,
all at the same time (see Figure~\ref{fig:inconsistent}). At the top, we observe
the case when nodes acquire the lock, one by one (in a sequential order in
the example) before attempting their task (updating or reading) and release the
lock afterwards. Because the second node acquires the lock after the first node
has finished updating, it reads the new parameter as $w_{2} = {[1,0,1]}^{\top}$.
At the bottom, we observe the case when individual coordinates of the parameter
vector are updated/read using atomic operations. In this case, the first two
nodes are accessing the first coordinate atomically at the same time (with a
load before store ordering in the example) whereas the third node is updating
the third coordinate. Later, the second node is reading the second coordinate
while the first one is updating the third coordinate. As a result, the second
node has an inconsistent view over the parameter vector (\ie, the vector, $w_{2}
= {[0,0,3]}^{\top}$, read by the second node would have never existed during the
update sequence). It is worth noting that, in this example, using mutex locks
takes more time (six units of time, discarding prefetching and caching effects)
than using atomic operations (roughly three units of time). In fact, this
observation is true when the updates are \emph{sparse}. Atomic operations
provide faster serialization (at the expense of inconsistent views) when fewer
nodes are competing for the same blocks of coordinates whereas mutex locks
provide more efficient serialization when the underlying operations are more
expensive (dense updates).

\emph{Distributed-memory architectures} cover networks of computers, IoT-enabled
edge devices and modern computer setups in which CPUs and accelerators work
together but have different physical memory spaces.
Figure~\ref{fig:architectures} (right) shows an example of a network of
computers in a particular communication topology. In this setup, also known as
the \emph{parameter server} setup~\cite{dean2012large,Li2013}, the communication is
\emph{centralized} around a set of nodes (called the \emph{masters} or the
\emph{servers}) that keep the shared parameters and constitute the hubs of a
star network. \emph{Worker} nodes (or \emph{clients}) pull the shared parameters
from and send their updates to the central nodes. 

When there are no masters in
the setup, and the nodes are allowed to communicate with each other in a more
general way, it becomes a \emph{multi-agent} setup. In this setup, updating the
shared parameter vector is \emph{decentralized} among the participating
\emph{agents} while obeying their respective communication topologies. The main
challenge in distributed-memory architectures, as opposed to shared-memory, is
that access to data in another node's memory space requires some sort of message
passing over network sockets. This, in turn, makes serialization and
synchronization operations rather expensive. Synchronous operations use blocking
communication primitives in the sense that all participating nodes have to wait
for a message before proceeding. Depending on the communication constraints and
topology, this can yield different communication complexities. For instance, in
a ring topology in which only point-to-point communication is allowed, and nodes
are queried in a round-robin fashion, we have $\bigO{n}$ communication
complexity. On the other hand, if many-to-many or all-to-all communication is
allowed, na\"{i}ve implementations offer $\bigO{n^{2}}$ complexities whereas
collective communication operations (such as broadcast and reduce) that use a
butterfly-like communication pattern achieve the optimal
$\bigO{\log\parentheses{n}}$~\cite{rabenseifner2004optimization}. Nevertheless, even the optimal synchronous
operations still suffer from the \emph{deadlock} (a situation in which all the
nodes are waiting for an output of each other or of a dead/offline node) and
\emph{straggler} (node that is slow due to either computation or communication
performance) problems, especially when messages are delivered slowly or may be lost altogether. 

To alleviate these problems, asynchronous operations are preferred,
although, this time, it becomes harder to design and analyze algorithms. In
asynchronous operations, nodes can interleave communication and computation
based on their own pace; yet, they have to deal with not only delayed
information over the parameter vector, when there is one writer and multiple
readers (\eg, one master node in a parameter server), but also inconsistent
views over the vector when the vector is shared among multiple masters.

\section{Centralized asynchronous algorithms}
\label{sec:shared}

Recall that our goal is to minimize an objective function of the form~\eqref{eq:erm}, and let us again suppose there is no regularizer, so that
\begin{align*}
f(w) = \frac{1}{m} \sum_{i=1}^m f_i(w).    
\end{align*}
In this section, we discuss asynchronous iterative methods for minimizing $f$ that involve updating a master copy of the optimization variables $w^{(k)}$.

The serial \sgd method discussed in \secref{sec:synchronous-sgd} performs updates of the form
\begin{align*}
w^{(k+1)} = w^{(k)} - \gamma^{(k)} s^{(k)}
\end{align*}
with search direction $s^{(k)} = \nabla f_{i^{(k)}}(w^{(k)})$ where $i^{(k)}$ is sampled uniformly at random between $1$ and $m$; \ie, the gradient of $f$ is approximated as the gradient at one of the data points. 

Although the search direction in \sgd is an unbiased estimator of $\nabla f$,
its variance (and higher moments) are typically non-zero, which limits the achievable solution accuracy. Specifically, if $f$ is $\mu$-strongly convex and each $f_i$ is $L$-smooth, then iterates of \sgd for a fixed step-size $\gamma^{(k)}=\gamma\in \left(0, \frac{1}{L}\right)$
satisfy
\begin{align*}
\mathbf{E}[\Vert w^{(k)}-w^{\star}\Vert^2] &\leq
\left[ 
1-2\gamma\mu(1-\gamma L)
\right]^k \Vert w^{(0)}-w^{\star} \Vert^2 \\
&\quad + \frac{\gamma\sigma^2}{\mu(1-\gamma L)},
\end{align*}
where  $w^\star = \arg\,\min_w f(w)$, and $\sigma^2$, defined as
\begin{align*}
\sigma^2=\frac{1}{m}\sum_{i=1}^m \left\Vert\nabla f_i(w^{\star})\right\Vert^2,
\end{align*}
is the variance of the search direction at the optimum~\cite[Theorem~2.1]{NWS:14}. The first term of the \sgd theoretical upper bound decays to zero at a linear rate, while the second term is a constant and describes the residual error. This means that \sgd with a fixed step-size $\gamma$ converges linearly
to a neighborhood of the optimum whose radius is proportional to $\sigma^2$ and $\gamma$. Decreasing $\gamma$ reduces the residual error, but it also results in a slower convergence. For any desired accuracy $\varepsilon>0$, letting
\begin{align*}
\gamma=\frac{\mu\varepsilon}{2\mu\varepsilon L + 2\sigma^2}
\end{align*}
ensures that $\mathbf{E}[\Vert w^{(k)}-w^{\star}\Vert^2]\leq \varepsilon$ for all iterations
\begin{align*}
    k\geq 2\left( Q + \frac{\sigma^2}{\mu^2\varepsilon}\right)\log(\varepsilon^0/\varepsilon),
\end{align*}
where $\varepsilon^0=\Vert w^{(0)}-w^{\star}\Vert^2$, and $Q=L/\mu$ is the condition number of the function $f$. This expression shows how a large value of~$\sigma^2$ forces a small step-size $\gamma$ to reach $\varepsilon$-accuracy, and therefore results in long convergence times.

It is common for the iteration complexity, and for optimal or allowable choices of algorithm parameters, to depend on the problem constants $L$ and $\mu$. In practice, these may not be directly known, and one may use an upper bound on $L$ and a lower bound on $\mu$ instead. An upper bound on the gradient Lipschitz constant $L$ can be easily obtained during the execution of an algorithm by tracking the ratio $\norm{\nabla f(x_{k+1}) - \nabla f(x_k)} / \norm{x_{k+1} - x_k}$. This leads to methods which resemble a backtracking line search, such as those described in~\cite{beck2009fast,schmidt2015nonuniform}, to upper bound the allowable step sizes. Note, however, that exactly tracking the aforementioned ratio requires the evaluation of the full gradient, which might not be plausible in the asynchronous setting. In such scenarios, techniques such as that in~\cite{hanzely2018accelerated} could be used to estimate upper bounds on the allowable step sizes \emph{without} requiring the full gradient evaluation. Estimating a lower bound on $\mu$ is much more challenging. One approach for estimating a lower bound on $\mu$ is provided in~\cite{nesterov2013gradient}. However, when using an $\ell_2$ regularizer of the form $r(w) = (\lambda /2 )\norm{w}_2^2$, then $\lambda$ directly provides a lower bound on $\mu$.

\subsection{Asynchronous parallel stochastic gradient methods} \label{sub-sec:async-sgd}

Recall that the mini-batch \sgd method, which uses the search direction~\eqref{eqn:mini-batch-direction}, can leverage multiple processors to compute gradient terms in parallel, but it is inherently synchronous: processors read the current decision vector, compute a gradient of their assigned component functions, and update the iterate. Once all $b$ gradient updates have been performed on the decision vector,
the algorithm proceeds to the next iteration.

As suggested in~\cite{Zinkevich:09}, this process can be pipelined. In such a realization, a master node interacts with worker nodes in a round-robin fashion, collecting their most recent stochastic (mini-batch) gradient, updating the decision vector and returning the new iterate to the worker before proceeding to serve the next worker. In a system with $n$ workers, each worker evaluates a stochastic gradient on a decision vector which is $\tau=n$ iterates old,
but can use a local mini-batch which is ${\mathcal O}(n)$ times larger and still finish its work before the next chance to interact with the master node. In this algorithm, the decision vector is thus updated using the search direction
\begin{align*}
s^{(k)} &= \frac{1}{b}\sum_{i\in I^{(k)}} \nabla f_i(w^{(k-\tau)}).
\end{align*}
Interestingly, the constant delay $\tau$ introduces negligible penalty in the convergence rate of the algorithm~\cite{Zinkevich:09}.

The round-robin interaction can be problematic in practice if some workers struggle to finish their work in time. The master may then have to wait, accept a suboptimal search direction from the worker, or skip its update altogether.
As shown in~\cite{AgD:11}, however, the same performance can be attained by an asynchronous parallel mini-batch algorithm which avoids global synchronization and allows worker nodes to read and write back to the master at their own pace.
In this case, the search direction used in the update rule of the algorithm is given by
\begin{align*}
s^{(k)} &= \frac{1}{b}\sum_{i\in I^{(k)}} \nabla f_i(w^{(k-\tau^{(k)})}),
\end{align*}
where $\tau^{(k)}$ is a time-varying delay capturing the staleness of the information used to compute the search direction for the $k^\textup{th}$ update.  In~\cite{FAJ:16}, it was shown that if $\tau^{(k)}$ is bounded, so that $\tau^{(k)}\leq \taubar$ for all $k$, then the iteration complexity of 
the asynchronous parallel mini-batch algorithm for strongly convex smooth optimization is given by
\begin{align*}
\mathcal{O}\left(\left( (\taubar+1)^2  Q+ \frac{\sigma^2}{\mu^2\varepsilon}\right)\log(1/\varepsilon)\right).
\end{align*}
In practice, $\taubar$ will depend on the number of parallel processors used for implementation of the algorithm. As long as $\taubar$ is of the order
$1/\sqrt{ \varepsilon}$, the iteration complexity of the asynchronous algorithm is asymptotically $\mathcal{O}((1/{\varepsilon}) \log(1/\varepsilon))$, which is exactly the iteration complexity achieved by serial~\sgd. This means that the delay becomes increasingly harmless as the asynchronous algorithm progresses. Furthermore, as $n$ workers are being run asynchronously and in parallel, updates may occur
roughly $n$ times as quickly, which means that a near-linear speedup in the number of workers can be expected.

When the decision vector dimension is very large, reading or writing the full vector takes considerable time. In a shared-memory system, it is then ineffective to lock the full vector during memory access. Instead, one typically only protects individual entries by using atomic read and write operations. 
This process is even more efficient if gradients are sparse and tend to have non-overlapping support. The probability that different workers simultaneously attempt to access the same elements of the decision vector is then low, and the $n$ workers effectively run independently in parallel. 

The first analysis of such a ``lock-free'' \sgd algorithm, called \hogwild, appeared in~\cite{RRW+:11}. To describe the convergence results, we need to introduce some additional notation. In particular, let $E_i$ be the support of $\nabla f_i(x)$ and define
\begin{align*}
\Delta &= \frac{\underset{j=1, \dots, d}{\max}\;\vert \{ i : j\in E_i \}\vert }{m}.
\end{align*}
The parameter $\Delta \in [\frac{1}{m},1]$ is a measure of the sparsity for the optimization problem.
In a fully dense dataset, $\Delta$ is equal to $1$ and in a completely sparse dataset, $\Delta$ is equal to $1/m$.

\hogwild lets each core run its own \sgd iterations without any attempt to coordinate or synchronize with other cores, repeating the following steps:
\begin{itemize}
\item Use atomic read operations to copy the shared decision variable $w$ into a local variable $\widehat{w}$.
\item Sample a component function $f_i$ and use $\widehat{w}$ to compute $s=\nabla f_i(\widehat{w})$. 
\item Use atomic write operations to update the current $w$ in shared memory
\begin{align*}
[w]_e \leftarrow [w]_e - \gamma [s]_e, \qquad \textup{for}\;\; e\in E_i.
\end{align*}
Note that for the write operation, only elements in the support of $\nabla f_i$ need to be updated.
\end{itemize}

During the execution of \hogwild, processors do not synchronize or follow an order between reads or writes. This implies that while one processor is evaluating its gradient, others may update the value of $w$ stored in the shared memory. Therefore, the value $\widehat{w}$ at which the gradient is calculated by a processor may differ from the value of $w$ to which the update is applied. Note also that a full vector $\widehat{w}$ read for a processor might not correspond to any state of $w$ in the shared memory at any time point (\cf\ Figure~\ref{fig:inconsistent}, bottom).

It was shown in~\cite{LPLJ:17} that the iteration complexity of \hogwild for smooth strongly convex optimization is
\begin{align*}
\mathcal{O}\left(\left( (\sqrt{\Delta}\taubar +1) Q+ \frac{\sigma^2}{\mu^2\varepsilon}\right)\log(1/\varepsilon)\right),
\end{align*}
where $\taubar$ is the maximum delay between reading and updating for cores. It follows that when $
\tau_{\max} = \mathcal{O}\left(\frac{1}{\sqrt{\Delta}}\right)$, \hogwild
converges at the same rate as the serial \sgd and therefore enjoys near-linear speedup. 

\subsection{Variance reduction and incremental aggregation methods}

A drawback with constant step-size \sgd algorithms, including the parallel and asynchronous variants discussed above, is that the iterates $\{w^{(k)}\}$  will not converge to an optimizer $w^{\star}$, but will exhibit a residual error. This error, which arises due to the mismatch between the gradients of individual component functions and their average, can be eliminated using incremental aggregation methods. Such methods maintain an estimate of the full gradient $\nabla f(w)$ whose error has a diminishing variance. 

Stochastic average gradient (\sag) is a randomized incremental aggregation method, which uses the search direction as the average of all component gradients evaluated at previous iterates. Specifically, at iteration $k$, \sag will have stored $\nabla f_i(w^{(d_i^k)})$ for all $i\in \{1,\ldots,m\}$, where $d_i^k$ represents the latest iterate at which $\nabla f_i$ was evaluated. An index $j\in\{1,\ldots,m\}$ is then drawn uniformly at random and the search direction is set by
\begin{align} \label{eqn:sag-search-direction}
s^{(k)} &= \frac{1}{m}\left(\nabla f_{j}(w^{(k)}) - \nabla f_j(w^{(d_j^k)})+\sum_{i=1}^m \nabla f_i(w^{(d_i^k)})\right) \nonumber\\
& = \frac{1}{m}\left(\nabla f_{j}(w^k) +\sum_{\mathclap{ i=1, i\neq j}}^m \nabla f_i(w^{(d_i^k)})\right).
\end{align} 
Although this $s^{(k)}$ is not an unbiased estimator of $\nabla f(w^{(k)})$, \sag enjoys a linear rate of convergence to the true optimizer without any residual error~\cite{Roux:12}. Specifically, \sag with the constant step-size $\gamma = \frac{1}{16 L}$ has the iteration complexity of
\begin{align*}
\mathcal{O}\left((Q + n)\log(1/\varepsilon)\right).  
\end{align*}

Inspired by \sag, several variance reduction methods were proposed including, to name a few, stochastic variance reduced gradient (\svrg)~\cite{Johnson:13}, stochastic average gradient with an unbiased estimator (\saga)~\cite{Defazio:14}, and stochastic dual coordinate accent (\sdca)~\cite{Shalev:13}.
The \sag method is a randomized variant of the incremental aggregated gradient method (\iag)~\cite{Blatt:07}. The search direction of \iag is identical to that of \sag \eqref{eqn:sag-search-direction}, but the index $j$ of the component function updated at every iteration is chosen cyclically rather than randomly. More precisely, the component functions are processed one-by-one using a deterministic cyclic order on the index set $\{1, 2,\ldots, m\}$, and hence, $d_i^k$ admits the recursion
\begin{align*}
  d_i^k = \begin{cases}
    k         & \text{if } i = (k-1\mod m) + 1 \,, \\
    d_i^{k-1} & \text{otherwise.}
  \end{cases}
\end{align*}

The natural parallelization strategy for \sag and \iag is the same as for \sgd: multiple workers draw independent indices uniformly at random from $\{1, \dots, m\}$, compute $\nabla f_i(x)$ and return these to a master that modifies the search direction, updates the decision vector, and pushes the updated decision vectors to idle workers. More precisely, the search direction used in the update rule of the asynchronous \iag is given by
\begin{align*}
s^{(k)} = \frac{1}{m}\sum_{i=1}^m f_i(w^{(d_i^k)}).
\end{align*} 
Note that the values 
\begin{align*}
\tau_i^k := k  - d_i^k, 
\end{align*}    
can be viewed as the delay encountered by the gradients of the component functions at $k^\textup{th}$ update. 

In~\cite{Mert:17}, it was shown that if $\tau_i^k\leq \tau_{\max}$ for all $i$ and $k\in\mathbb{N}$, then asynchronous \iag with constant step-size 
\begin{align*}
\gamma \in \left(0,\;  \frac{8\mu}{25 L (\tau_{\max}+1)(\mu + L)}\right)    
\end{align*}
requires 
\begin{align*}
\mathcal{O}\left((\tau_{\text{max}}+1)^2  Q^2\log(1/\varepsilon)\right)
\end{align*}
iterations to achieve an $\varepsilon$-optimal solution. Since the analysis in~\cite{Mert:17} considers deterministic guarantees on $\varepsilon$-optimality, it is natural that the convergence time bounds are more conservative than those of its stochastic counterparts.

In~\cite{LPLJ:17}, a lock-free asynchronous
version of \saga, called \asaga, was proposed. If the maximum delay bound in the \asaga implementation satisfies $\tau_{\max} < m/10$, then the iteration complexity is given by
\begin{align*}
\mathcal{O}\left(\left((\sqrt{\Delta}\tau_{\max} +1)Q + m\right)\log(1/\varepsilon)\right).  
\end{align*}
Therefore, \asaga obtains the same iteration complexity as \sag and \saga when
$\tau_{\max}$ satisfies $\tau_{\max} \leq \mathcal{O}(m)$ and
\begin{align*}
\tau_{\max} \leq \mathcal{O}\left( \frac{1}{\sqrt{\Delta}}\max\left\{1,\frac{m}{Q}\right\}\right).
\end{align*}
This means that in the well-conditioned regime where $m > Q$, a linear speedup is theoretically possible for \asaga even without sparsity. This is in contrast to some work on asynchronous incremental gradient methods
which required sparsity to get a theoretical linear speedup over their sequential counterpart~\cite{MPP+:17}.

\subsection{Asynchronous coordinate descent methods}
Stochastic gradient methods handle datasets with many samples $m$ by using a search direction that avoids evaluating the gradient of the loss at every sample. Similarly, \emph{coordinate descent} methods address problems with large decision vector dimension $d$ by avoiding to compute updates for every decision variable in every iteration. 

Coordinate descent methods, which traditionally cycle through coordinates in a deterministic order, have a long history in optimization (see, \eg,~\cite{LuT:92}). The research was revitalized by Nesterov's elegant analysis of randomized coordinate descent methods~\cite{Nes:12}. At each iteration $k$, these methods draw a random coordinate $j(k)$ from $\{1, \dots, d\}$ and perform the update
\begin{align}
    [w^{(k+1)}]_{j(k)} &= [w^{(k)}]_{j(k)} - \gamma [\nabla f(w^{(k)})]_{j(k)} \label{eqn:coordinate_descent_iteration}
\end{align}
Similarly to mini-batching in \sgd, one is not restricted to picking a single coordinate to update in each iteration, but can sample random subsets (blocks) of coordinates~\cite{Nes:12}. 

Synchronous parallel coordinate descent methods have been suggested in, \eg,~\cite{BKB:11, RiT:16}. In each iteration of these methods, a master node draws $n$ (blocks of) coordinates and distributes the work to evaluate the associated partial gradients on $n$ workers. The master waits for all workers to return before it updates the decision vector and continues with the next iteration. An asynchronous coordinate descent method for shared-memory architectures was proposed and analyzed in~\cite{LiW:15}. In essence, this method spawns $n$ parallel coordinate descent processes. In each process, a worker thread performs an inconsistent read of $w^{(k)}$ from the shared memory, draws a random coordinate index and performs the update (\ref{eqn:coordinate_descent_iteration}) using atomic writes. Linear convergence is proven under the assumption that the maximum overlap $\tau_{\max}$ (defined in the same way as for \hogwild above) is small enough. When the coupling between components is weak (in a precise sense defined in~\cite{LiW:15}), $\tau_{\max}$ can be of order $d^{1/4}$, while the maximal admissible $\tau_{\max}$ shrinks close to one when the coupling is strong. An extension of this asynchronous coordinate scheme to operator mappings is presented in~\cite{ZYM+:16}.

Note that coordinate descent methods are naturally variance-reduced, since partial derivatives at the optimum are all zero, \textit{i.e.}, $[\nabla f(w^\star)]_j=0$ for all $j=1,\ldots,d$. 
Thus, the value of variance-reduced coordinate descent methods may appear limited. However, the main drawback of many coordinate descent methods is that they cannot handle non-separable regularizers. Variance-reduced coordinate descent methods, on the other hand, allow us to solve optimization problems with arbitrary (not necessarily separable) regularizers~\cite{Hanzely:18,Hanzely:20}.

\subsection{Proximal methods for convex and non-convex optimization}
For ease of exposition, we have described stochastic gradient methods for smooth and strongly convex losses. However, many of the results extend directly to proximal gradient methods for optimization problems with convex and non-convex loss functions plus a possibly non-smooth regularization term. Specifically, the results in~\cite{FAJ:16,LiW:15} already consider composite optimization problems comprising a smooth finite-sum term and a non-smooth regularizer. An extension of \asaga to such problems is described and analyzed in~\cite{PLL:17}. Convergence rate of asynchronous mini-batch algorithms and randomized coordinate descent methods for non-convex optimization are studied in~\cite{Lian:15}. Extensions of \hogwild to non-convex optimization problems are presented in~\cite{De:15} and \cite{Nguyen:18}. In~\cite{Tseng:04}, a theoretical upper-bound on the convergence rate of \iag for non-convex composite optimization is derived.

\subsection{Analysis techniques}
As mentioned above, the framework of~\cite{Bertsekas:89} for partially
asynchronous algorithms (\ie, those with bounded delay), is not directly
applicable to stochastic optimization methods described above. Instead, convergence guarantees have typically been established on a per-algorithm basis, often using complex and laborious induction proofs in which sources of conservatism are hard to isolate. A closer analysis of these proofs reveals that they rely on a few common principles. One such principle is to introduce a well-defined global ordering of events in the system, and model (bounded) asynchrony as (bounded) time-varying delays~\cite{Bertsekas:89}. As discussed in~\cite{LPLJ:17}, this ordering may be non-trivial in algorithms such as \asaga and \hogwild. Another principle is to view iterates as perturbed versions of ideal quantities \cite{MPP+:17}. A third principle is to reduce expressions for the evolution of the iterate suboptimality, which typically depends on many previous iterates, to standard forms that are well-understood. A number of such sequence results, derived specifically for asynchronous optimization algorithms, are introduced in~\cite{FAJ:14,AFJ:16}.

To be more concrete, let us consider asynchronous algorithms as gradient iterations with additive gradient errors, \ie,
\begin{align*}
  w^{(k+1)} = w^{(k)} - \gamma \bigl(\nabla f(w^{(k)}) + e^{(k)}\bigr).
\end{align*}
Similar to known lines of convergence proofs for gradient and subgradient methods with errors, it follows
directly by expanding the squared norm of the iterate error that
\begin{align*}
  {||w^{(k+1)} - w^{\star}||}^{2}
  & = {||w^{(k)} - w^{\star}||}^{2} - 2\gamma{\langle w^{(k)} -
    w^{\star}, \nabla f(w^{(k)})\rangle} \\
  & \quad {} + \gamma^2 ||\nabla f(w^{(k)})||^2 + E^{(k)},
\end{align*}
where the gradient errors are encapsulated by the last term
\begin{align*}
  E^{(k)} = \gamma^2 {||e^{(k)}||}^2 - 2 \gamma \langle w^{(k)} - \gamma
    \nabla f(w^{(k)}) - w^\star, e^{(k)} \rangle.
\end{align*}
Letting $V^{(k)}= {||w^{(k)} - w^\star||}^2$, and using standard strong
convexity and smoothness inequalities~\cite{nesterov2004introductory} allow us to derive
\begin{align*}
  V^{(k+1)} \leq \left(1 - 2\gamma \frac{\mu L}{L + \mu}\right) V^{(k)}  +
    E^{(k)},
\end{align*}
for any step-size $\gamma \in \left(0, \frac{2}{L + \mu}\right]$. Note that when
$E^{(k)} = 0$, taking $\gamma = \frac{2}{L + \mu}$ leads to
\begin{align*}
  V^{(k+1)} &\leq \left(\frac{Q -1}{Q + 1}\right)^2 V^{(k)},
\end{align*}
which guarantees linear convergence of the iterates to the optimum. This is the standard analysis of the gradient descent method. 
For the asynchronous case, the error term $E^{(k)}$ can often be bounded by terms involving only distances of the current and past iterates from $w^\star$, \ie,
\begin{align*}
  V^{(k+1)} \leq \left(1 - 2\gamma \frac{\mu L}{L + \mu}\right) V^{(k)}  +
    {\mathcal H}\bigl(V^{(k)}, V^{(k-1)},\ldots,V^{(0)}\bigr),
\end{align*}
where the function ${\mathcal H}$ models the history dependence. For example, the analysis of \iag in~\cite{Mert:17} establishes the bound
\begin{align*}
  {\mathcal H}\bigl(V^{(k)}, V^{(k-1)},\ldots,V^{(0)}\bigr) = (6\gamma^2 L^2\tau
    + 9\gamma^4 L^4 \tau^2) \max_{\mathclap{k - 2\tau\leq s \leq k}} V^{(s)}.
\end{align*}
To analyze the effect of the asynchrony on \iag, it is then convenient to use the following sequence result.

\begin{lemma}[\cite{FAJ:14}]
  Let $\{V^{(k)}\}$ be a nonnegative sequence satisfying
\begin{align*}
  V^{(k+1)} \leq p V^{(k)} + q  \max_{k - d\leq s \leq k} V^{(s)}, \quad k \in \mathbb{N},
\end{align*}
for some positive integer $d$ and non-negative scalars $p$ and $q$ such that $p + q \leq 1$. Then, we have
\begin{align*}
  V^{(k)} \leq \rho^k V^{(0)}, \quad k \in \mathbb{N},
\end{align*}
where $\rho = (p + q)^{\frac{1}{p + q}}$.
\end{lemma}

Using this result with 
\begin{align*}
d=2\tau, \quad p = 1 - 2\gamma \frac{\mu L}{L + \mu}, \quad q = 6\gamma^2 L^2 \tau + 9\gamma^4 L^4 \tau^2,
\end{align*}
yields that the iterates generated by \textsc{Iag} with constant step-size
\begin{align*}
\gamma \in \left(0, \frac{8\mu}{25 L (\tau+1)(\mu + L)}\right)
\end{align*}
are globally linearly convergent~\cite{Mert:17}.

For stochastic asynchronous algorithms, it is sometimes more natural to interpret the algorithmic effects of asynchrony as perturbing the stochastic iterates with bounded noise~\cite{MPP+:17}. Consider the following iteration
\begin{align*}
  w^{(k+1)} = w^{(k)} - \gamma g(w^{(k)} + \eta^{(k)}, \xi^{(k)}),
\end{align*}
where $\eta^{(k)}$ is a stochastic error term, $\xi^{(k)}$ is a random variable
independent of $w^{(k)}$, and $g$ is an unbiased estimator
of the true gradient of $f$ at $w^{(k)}$:
\begin{align*}
  \mathbb{E}_{\xi^{(k)}} \bigl[g(w^{(k)}, \xi^{(k)})\bigr] = \nabla f(w^{(k)}).
\end{align*}
Let $\widehat{w}^{(k)} = w^{(k)} + \eta^{(k)}$. Then,
\begin{align*}
  ||w^{(k+1)} - w^\star||^2
    & = ||w^{(k)} - w^\star||^2  + \gamma^2 ||g(\widehat{w}^{(k)})||^2 \\
    & \quad {} - 2\gamma \langle \widehat{w}^{(k)} - w^\star,
      g(\widehat{w}^{(k)})\rangle \\
    & \quad {} + 2\gamma \langle \widehat{w}^{(k)} - w^{(k)},
      g(\widehat{w}^{(k)})\rangle.
\end{align*}
Assume that $\widehat{w}^{(k)}$ and $\xi^{(k)}$ are independent. Then, taking expectation and using a standard strong convexity bound as well as a squared triangle inequality yields
\begin{align*}
  V^{(k+1)}
    & \leq \left(1 - \frac{\gamma \mu}{2}\right) V^{(k)} - 2\gamma X^{(k)} +
      \gamma^2 \underbrace{\mathbb{E}[||g(\widehat{w}^{(k)})||^2]}_{R_0^{(k)}}
      \\
    & \quad {} + \gamma \mu \underbrace{\mathbb{E}[|| \widehat{w}^{(k)} -
      w^{(k)}||^2]}_{R_1^{(k)}} \\
    & \quad {} + 2\gamma \underbrace{\mathbb{E}[\langle  \widehat{w}^{(k)} -
      w^{(k)}, g(\widehat{w}^{(k)})\rangle]}_{R_2^{(k)}},
\end{align*}
where $V^{(k)} =  \mathbb{E}[||w^{(k)} - w^\star||^2]$ and $X^{(k)} =
\mathbb{E}[f(w^{(k)}) - f^\star]$. Note that $R_0^{(k)}$, $R_1^{(k)}$, and
$R_2^{(k)}$ are error terms due to asynchrony: 
$R_0^{(k)}$ captures the delayed gradient decay with each iteration, $R_1^{(k)}$
represents the mismatch between the true iterate and its noisy (outdated)
estimate, and $R_2^{(k)}$ measures the size of the projection of that mismatch
on the gradient at each step. To derive the convergence rate, we bound these
error terms using past values of $X^{(k)}$, \ie,
\begin{align*}
  V^{(k+1)}
    & \leq \left(1 - \frac{\gamma \mu}{2}\right) V^{(k)} - 2\gamma X^{(k)} \\
    & \quad {} + {\mathcal H}\bigl(X^{(k)}, X^{(k-1)}, \ldots, X^{(0)}\bigr).
\end{align*}
For example, \textsc{Hogwild!} admits the history function
\begin{align*}
  {\mathcal H}\bigr(X^{(k)}, X^{(k-1)}, \ldots, X^{(0)}\bigr)
    & = 4\gamma^2 L C_2 \sum_{j=k - \tau}^{k-1} X^{(j)} \\
    & \quad {} + 4 \gamma^2 L C_1 X^{(k)},
\end{align*}
where $C_1 = 1 + \sqrt{\Delta} \tau$ and $C_2 = \sqrt{\Delta} + \gamma \mu C_1$~\cite{LPLJ:17}. The following result is then convenient to apply.

\begin{lemma}[\cite{AFJ:16}]
  Assume the non-negative sequences $\{V^{(k)}\}$ and $\{X^{(k)}\}$ satisfy 
\begin{align}
  V^{(k+1)} \leq a V^{(k)} - b X^{(k)} + c \sum_{j=k-\tau}^{k} X^{(j)},\label{eqn:sequence}
\end{align}
where $a \in (0,1)$, and $b$ and $c$ are nonnegative real numbers. If $a<1$ and 
\begin{align*}
\frac{c}{1-a} \frac{1-a^{\tau+1}}{a^{\tau}} \leq b,
\end{align*}
then $V^{(k)} \leq a^k V^{(0)}$ for all $k\in \mathbb{N}_0$.
\end{lemma}

Using this result with $a = 1-\gamma \mu/2$, $b = 2\gamma$, and $c = 4\gamma^2 L C_2$, immediately yields the convergence rate of \textsc{Hogwild!} stated in Section~\ref{sub-sec:async-sgd} above.

A similar analysis can be made for \iag, \textsc{Hogwild!}, and many other algorithms, also in the absence of strong convexity.  These results require slightly different sequence results. It is also possible to derive convergence rate results for iterations with unbounded delays. In fact,~\cite{Feyzmahdavian:14} presents the convergence rate results for asynchronous max-norm contractions for both the totally and partially asynchronous models.

There are methods that use an aggregate of iterates instead of the current iterate in their update rules, such as variance reduction methods \miso~\cite{Mairal:15} and \finito~\cite{defazio2014finito}, incremental gradient methods~\cite{Mokhtari:18}, and delay-tolerant gradient methods~\cite{Mishchenko:18}. To be more specific, let $y_i^{(k)}$ be the copy of the decision variable $w$ used in the most recent computation of $\nabla f_i$ available at iteration $k$. The variable $y_i^{(k)}$ is updated as
\begin{align*}
  y_i^{(k)} = \begin{cases}
    w^{(k)}         & \text{if } i = i(k) \,, \\
    y_i^{(k-1)} & \text{otherwise,}
  \end{cases}
\end{align*}
where $i(k)$ is the index of the component function chosen uniformly at random at step $k$. Then, the update of \miso and \finito is given by
\begin{align*}
    w^{(k+1)} = \frac{1}{m}\sum_{i=1}^m y_i^{(k)} - \gamma \sum_{i=1}^m \nabla f_i(y_i^{(k)}).
\end{align*}
Since the update rule of these algorithms cannot be written as 
\begin{align*}
    w^{(k+1)} = w^{(k)} - \gamma s^{(k)},
\end{align*}
their convergence does not follow directly from the arguments above. Nevertheless, we note that the proof in~\cite[Equation 8]{Mishchenko:18} hinges on the same argument as Lemma~1.

\section{Decentralized algorithms}
\label{sec:multi-agent}

Decentralized algorithms (also known as ``consensus,'' ``gossip,'' or
``multi-agent'' algorithms) are an alternative to parameter-server algorithms in
the distributed memory setting. As the name suggests, in decentralized
methods, there is no authoritative copy of the parameters; rather, each worker maintains and updates a local working copy of the optimization variables.
In contrast to centralized methods, decentralized methods do not have a single bottleneck or point-of-failure, and thus may potentially scale to larger problems.

Consider a system with $n$ workers, and let $w_i^{(k)}$ denote the copy at worker $i$ after $k$ iterations. A simple synchronous decentralized method starts with all nodes at the same initial point $w_i^{(0)} = w^{(0)}$ and repeats updates of the form
\begin{equation} \label{eq:synchronous-decentralized}
w_i^{(k+1)} = \sum_{j=1}^n P_{i,j}^{(k)} w_j^{(k)} - \gamma^{(k)}_i s_i^{(k)},
\end{equation}
where $P^{(k)}_{i,j}$ is a scalar between $0$ and $1$ quantifying the influence of worker $j$ on worker $i$, and $s_i^{(k)}$ is the search direction computed at worker $i$. 

If $P_{i,j}^{(k)} = 1/n$ for all $i,j \in [n]$ and $k \ge 0$, then the variables at every worker are identical (\ie, exactly equal to their average) after every update. Furthermore, if the directions $s_j^{(k)}$ are independent stochastic gradients computed using $b$ gradient samples, then the method is equivalent to \sgd with mini-batch size $n b$. Implementing this update with $P_{i,j}^{(k)} = 1/n$ requires coordination among all workers. This can be accomplished in a communication-efficient way using the \textsc{AllReduce} primitive mentioned in \secref{sec:architectures}; we refer to such a method as \textsc{AllReduce} \sgd (\textsc{ar-sgd})~\cite{Agarwal2014reliable,chen2016revisiting}. In a system where $P_{i,j}^{(k)} = 1/n$ for all $i,j \in [n]$ and $k \ge 0$, each node could viewed as being a worker \emph{and} a central authority for the purpose of contrasting with the methods discussed in \secref{sec:shared}.

In general, the updates at each worker need not depend on the values from all other workers. For more general values of $P_{i,j}^{(k)}$, worker $i$ only needs to receive messages from worker $j$ if $P_{i,j}^{(k)} > 0$. When there are $n$ workers, the entries $P_{i,j}^{(k)}$ can be collectively viewed as an $n \times n$ matrix $P^{(k)}$. Equivalently, one can form a communication graph with one vertex for each worker and with an edge set $\mathcal{E}^{(k)}$ containing a (directed) edge from $j$ to $i$ if $P_{i,j}^{(k)} \ne 0$.

A natural way to generalize the notion of averaging while reducing the
communication overhead is to make use of matrices $P^{(k)}$ (equivalently, communication graphs) that are sparse, and that correspond to a diffusion or
random walk. In an asynchronous implementation, it is possible that messages may
be delayed, and so it is not practical to assume that $P^{(k)}$ is symmetric and
doubly-stochastic,\footnote{A matrix $P$ is row-stochastic (respectively,
column-stochastic) if each row (respectively, column) of $P$ sums to 1, and
all entries are non-negative. $P$ is doubly-stochastic if it is both row- and
column-stochastic.} since this would impose that if $i$ receives a message
from $j$ then $j$ also receives one from $i$ to perform an update.  Such a
method is referred to as \emph{push-pull} since it requires two-way exchange of
information.  Doubly-stochastic methods guarantee that workers converge to the
network-wide average at a geometric rate, but they are inherently synchronous.  

Methods using row-stochastic $P^{(k)}$ are referred to as \emph{pull-based} methods; they only involve a one-way exchange of information, and the weights $P^{(k)}_{i,j}$ are determined and applied by the receiver $i$. Row-stochastic methods guarantee that the vectors $w^{(k)}_i$ at each worker converge to a consensus at a geometric rate. However, the consensus values are not necessarily an unbiased estimate of the network-wide average; this bias in the consensus values can prevent the iterates in~\eqref{eq:synchronous-decentralized} from converging to a minimizer of~\eqref{eq:erm}.

Methods using column-stochastic $P^{(k)}$ are referred to as \emph{push-based}
methods; they only involve a one-way exchange of information, and the
weights $P^{(k)}_{i,j}$ are determined and applied by the sender $j$.
Column-stochastic methods guarantee that the vectors $w^{(k)}_i$ at each worker
converge at a geometric rate, but not necessarily to a consensus. Rather, the limit value depends on the message-passing topology
(through the stationary distribution, if $P^{(k)} = P$ is constant over time and
seen as the transition matrix of a Markov chain). However, unlike row-stochastic
methods, this discrepancy is easy to correct in column-stochastic methods. 

The \pushsum algorithm~\cite{kempe2003gossip} (also called \emph{ratio
consensus}) is a column-stochastic method in which each worker tracks one
additional parameter, referred to as the push-sum weight, that can be used to
compensate for discrepancies due to the message-passing topology. The push-sum weight, which we denote by $\phi_i^{(k)}$, is
initialized to $1$ at every worker. Whenever a worker communicates its
parameters, it also communicates the push-sum weight. Any imbalance built up in
the parameters also appears in the push-sum weight. Therefore, by rescaling the
parameters by the push-sum weight, workers running the \pushsum
algorithm are guaranteed to converge to a consensus on the network-wide average
at a geometric rate. Decentralized optimization methods built on
\pushsum have the form,
\begin{align}
\label{eq:synchronous-col-decentralized}
\begin{split}
    w_i^{(k+1)} &= \sum_{j=1}^n P_{i,j}^{(k)} w_j^{(k)} - \gamma^{(k)}_i s_i^{(k)},\\
    \phi_i^{(k+1)} &= \sum_{j=1}^n P_{i,j}^{(k)} \phi_j^{(k)},\\
    z_i^{(k+1)} &= \frac{w_i^{(k+1)}}{\phi_i^{(k + 1)}}.
\end{split}
\end{align}

The (synchronous) Stochastic Gradient-Push (\sgp) algorithm~\cite{nedic2016stochastic} is an analog of stochastic gradient descent for decentralized optimization.
Specifically, \sgp uses updates~\eqref{eq:synchronous-col-decentralized} where the search direction $s_j^{(k)}$ is a stochastic mini-batch gradient evaluated by worker $j$ at the rescaled point $z_j^{(k)}$ on a subset of the data $I^{(k)}_j$,
\[
    s_j^{(k)} = \sum_{m \in I^{(k)}_j} \nabla \ell \left(p(x_m; z^{(k)}_j),\ y_m\right).
\]
When the functions $f_i$ are strongly convex, each worker $j$ running \sgp with a diminishing step-size is guaranteed to converge to a minimizer of $f$~\cite{nedic2016stochastic}:
\[
    f(\hat{z}^{(K)}_j) - f(w^\star) \leq \mathcal{O}\left(\frac{\log{K}}{K}\right),
\]
where $\hat{z}^{(K)}_j$ is a weighted average of the sequence $\{ z^{(k)}_j \}^K_{k=0}$ produced at worker $j$.

The \sgp algorithm is synchronous since each worker $i$ blocks to send and receive messages from other workers $j$ for which $P_{i,j}^{(k)} > 0$ before proceeding to the next iteration.
Since \emph{pull-based} methods and \emph{push-based} methods only involve a one-way exchange of information, they are readily amenable to asynchronous implementations.
In the next section we will describe some specific asynchronous decentralized optimization methods along with their known convergence guarantees.

\subsection{Asynchronous decentralized methods}
\label{sub-sec:decentralized-methods}

The Overlap Stochastic Gradient-Push (\osgp) algorithm~\cite{assran2019stochastic} builds on \sgp by allowing for message delays. \osgp uses the same search direction as \sgp, but reduces the communication and synchronization overhead by overlapping communication of parameters between workers with multiple stochastic gradient updates. Let $\tau^{(k)}_{i,j}$ denote the delay experienced by a message sent from worker $j$ and received by worker $i$ at iteration $k$ (\ie, the message was transmitted at time $k - \tau^{(k)}_{i,j}$). By convention, we take $\tau^{(k)}_{i,i} = 0$ for all $k$ and $i\in[n]$. Let $\mathcal{M}^{(k)}_i$ denote the set such that $\tau_{i,j}^{(k)} \in \mathcal{M}^{(k)}$ implies that worker $i$ received a message from $j$ at iteration $k$ with delay $\tau^{(k)}_{i,j}$. The updates in \osgp can be written in terms of these delayed indices as
\begin{align}
\label{eq:osgp}
\begin{split}
    w_i^{(k+1)} &= \sum_{\tau^{(k)}_{i,j} \in \mathcal{M}^{(k)}_i} P_{i,j}^{(k - \tau^{(k)}_{i,j})} w_j^{(k - \tau^{(k)}_{i,j})} - \gamma^{(k)}_i s_i^{(k)},\\
    \phi_i^{(k+1)} &= \sum_{\tau^{(k)}_{i,j} \in \mathcal{M}^{(k)}_i} P_{i,j}^{(k - \tau^{(k)}_{i,j})} \phi_j^{(k - \tau^{(k)}_{i,j})}.
\end{split}
\end{align}
The rescaling update for $z^{(k)}_i$ is identical to the one in \sgp.
Note that, although \osgp handles message delays, it does not deal with computation delays (\ie, heterogeneous update rates amongst workers).
Specifically, each \osgp worker $i$ must perform the updates
in~\eqref{eq:osgp} at every iteration.

The Asynchronous Gradient-Push algorithm (\agp) proposed in~\cite{assran2017empirical} and analyzed in~\cite{assran2018asynchronous} is an analog of gradient descent for asynchronous decentralized optimization, and deals with both message and computation delays.
This algorithm is similar to \sgp, but removes all synchronization points.
Let $\delta^{(k)}_i \in \{0, 1\}$ denote a binary indicator that is equal to $1$ if worker $i$ completes an update at iteration $k$, and is equal to $0$ otherwise. The global iteration counter $k$ (used only to describe the algorithm) increments whenever any worker (or subset of workers) completes a gradient-based update. If worker $i$ completes an update at iteration $k$ (\ie, $\delta^{(k)}_i=1$), then the \agp update is identical to that in~\eqref{eq:osgp}. If worker $i$ does \emph{not} complete an update at iteration $k$ (\ie, $\delta^{(k)}_i = 0$), then its iterates remain unchanged
\begin{align}
w^{(k+1)}_i = w^{(k)}_i, \quad \phi^{(k+1)}_i = \phi^{(k)}_i, \quad z^{(k+1)}_i = z^{(k)}_i.
\end{align}
In contrast to \osgp, note that the \agp workers do not necessarily update their parameters at every iteration.

Suppose the message delays and the time between an \agp worker's successive updates are bounded. 
When the functions $f_i$ are strongly convex and $L$-smooth, workers running \agp up to a global iteration $K$ minimize a re-weighted version of~\eqref{eq:erm}, defined as~\cite{assran2018asynchronous}
\begin{align}\label{eq:rw-prob}
  \begin{aligned}
    & \minimize_{w \in \R^d} & & \sum_{i=1}^n \overline{p}^{(K)}_i f_i(w) \,,
  \end{aligned}
\end{align}
where the re-weighting values $\overline{p}^{(K)}_i > 0$ are given by
\begin{equation}
	p^{(K)}_i \coloneqq \sum^{K - 1}_{k=0} \gamma^{(k)}_i \delta^{(k)}_i, \quad \text{and} \quad \overline{p}^{(K)}_i \coloneqq \frac{p^{(K)}_i}{ \sum^n_{i=1} p^{(K)}_i }.
\end{equation}
In particular, letting $w^\star_K$ denote the minimizer of~\eqref{eq:rw-prob}, it can be shown that~\cite[Theorems 4 \& 5]{assran2018asynchronous}
\[
    \frac{1}{K}\sum^K_{k=0} \norm{\frac{1}{n}\sum^n_{i=1}z^{(k)}_i - w^\star_K}^2 \leq \mathcal{O}\left(\frac{1}{\sqrt{K}}\right).
\]
If all workers use the same constant step-size and perform a similar number of gradient-based updates by the end of training, then $\overline{p}^{(K)}_i \approx 1/n$, and the workers converge to the unbiased minimizer of the objective in~\eqref{eq:erm}. On the other hand, workers that perform more updates than their peers bias the solution towards their local objective. 

This convergence theory also suggests an approach for correcting the bias: slower workers can use larger step-sizes in order to compensate for their slower update rates. To do this workers need an idea of how many updates they have performed locally relative to the total number of updates performed by all workers. This can also be estimated in a decentralized way by communicating one additional scalar variable, and when the functions $f_i$ are convex and smooth, it can be shown that~\cite{zhang2019asyspa} 
\[
    \max_{k \leq K} f(z^{(k)}_j) - f_i(w^\star) \leq \mathcal{O}\left(\frac{\log{K}}{\sqrt{K}}\right).
\]
In~\cite{olshevsky2018robust} a similar method is analyzed in the context of stochastic gradients. 

Further improvements are obtained in~\cite{tian2018asy} and~\cite{tian2019asynchronous} by incorporating robust \pushsum, which tolerates dropped messages by using additional memory at each worker, and by incorporating gradient-tracking schemes, which lead to faster iteration-wise convergence but also involve twice the communication overhead per iteration, and hence may not be practical for machine learning problems with high-dimensional models.

We note that decentralized asynchronous methods have also been proposed based on applying coordinate descent methods to a dual formulation~\cite{bianchi2015coordinate,lu2015dual}. However, while these methods allow for randomized update order, they are not asynchronous in the sense considered in this paper, of allowing for communication and computation delays.

In the next subsection, we will describe general proof techniques for analyzing asynchronous decentralized optimization algorithms under bounded message \emph{and} computation delays. Following that discussion, we will summarize empirical assessments of these methods in the literature, and conclude by describing practical challenges and open problems in this budding research area.

\subsection{Analysis}
\label{sub-sec:anal}

Similar to centralized methods, analysis techniques for decentralized methods also exploit the idea of using a well-defined order of events in the system. However, decentralized methods require fundamentally different analysis techniques than centralized methods. To simplify the discussion, suppose that the averaging weights $P^{(k)}_{i,j}$ are static (i.e., workers always choose the same averaging weights to communicate with their neighbours).
Equations such as~\eqref{eq:synchronous-decentralized} and~\eqref{eq:synchronous-col-decentralized} describe synchronous decentralized optimization algorithms from an individual worker's perspective.
However, it is typically easier to study these methods from a global perspective by collectively viewing the entries $P_{i,j}$ as an $n \times n$ matrix $P$ and viewing the variables $w^{(k)}_i$ and $s^{(k)}_i$ as the rows of $n \times d$ matrices $W^{(k)}$ and $S^{(k)}$ respectively. Then equation~\eqref{eq:synchronous-decentralized} can be re-written in matrix-vector form as
\begin{equation}
    \label{eq:glob-synchronous-decentralized}
    W^{(k + 1)} = P W^{(k)} - \Gamma^{(k)} S^{(k)},
\end{equation}
where $\Gamma^{(k)}$ is an $n \times n$ diagonal matrix with the step-size $\gamma^{(k)}_i$ on the $i^{th}$ diagonal.
Similarly,~\eqref{eq:synchronous-col-decentralized} can be re-written as
\begin{align}
\label{eq:glob-synchronous-col-decentralized}
\begin{split}
    W^{(k + 1)} &= P W^{(k)} - \Gamma^{(k)} S^{(k)} \\
    \phi^{(k+1)} &= P \phi^{(k)} \\
    Z^{(k+1)} &= \text{diag}(\phi^{(k+1)})^{-1} W^{(k+1)},
\end{split}
\end{align}
where $\phi^{(k+1)}$ is an $n \times 1$ vector containing the push-sum weights, and $\text{diag}(\phi^{(k+1)})$ is a diagonal matrix with the push-sum weight $\phi^{(k+1)}_i$ on the $i^{th}$ diagonal.

Broadly, there are three main steps involved in proving convergence of workers' parameters under the assumption of bounded message and computation delays: (i) mathematically modelling delays, (ii) proving convergence of the optimization iterates to a consensus sequence under the delay model, (iii) proving convergence of the consensus sequence to a minimizer.

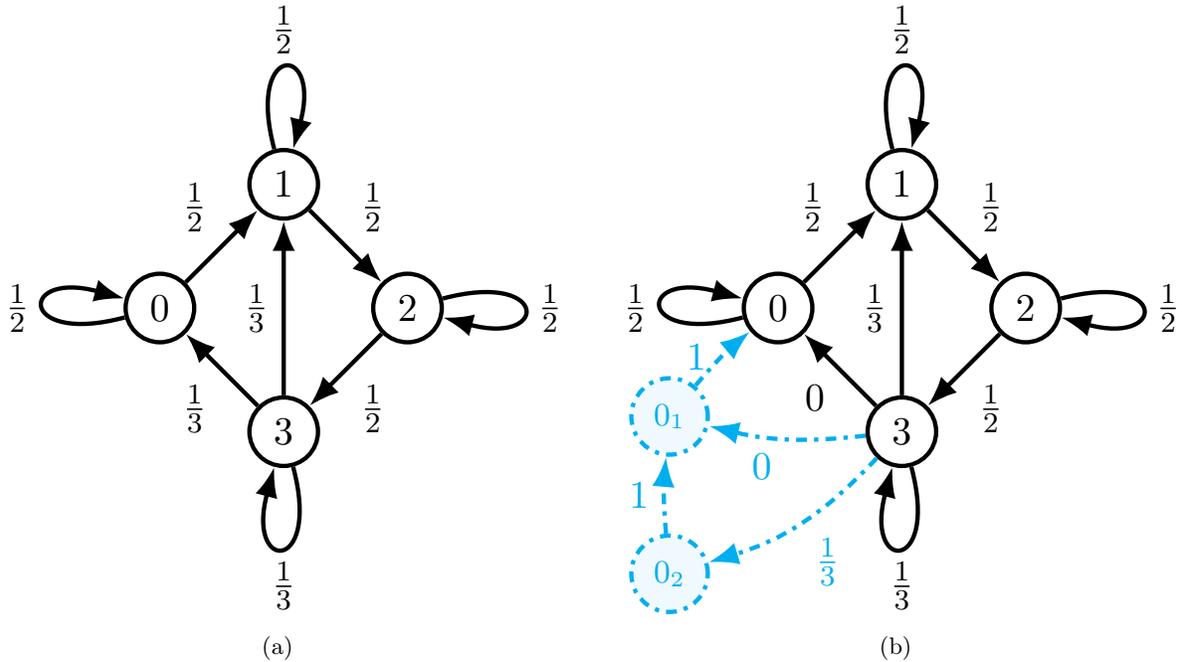
\begin{figure}[t]
\centering
{
\tikzstyle{edge} = [very thick, -{Latex}]
\tikzstyle{vedge} = [very thick, cyan, dashdotted, -{Latex}]
\tikzstyle{non-virtualnode} = [draw, circle, fill=white, very thick]
\tikzstyle{virtualnode} = [draw, circle, cyan, fill=cyan!5, very thick, dashdotted]
\subfloat[]{
\resizebox{0.49\columnwidth}{!}{
\begin{tikzpicture}[baseline]
% Nodes
\node[non-virtualnode] (topcircle) {$1$};
\node[non-virtualnode] (leftcircle) [below left = 1cm of topcircle] {$0$};
\node[non-virtualnode] (rightcircle) [below right = 1cm of topcircle] {$2$};
\node[non-virtualnode] (bottomcircle) [below right = 1cm of leftcircle] {$3$};
% Lines
% --
\draw[edge] (leftcircle) -- node[above left]{$\frac{1}{2}$} (topcircle);
\draw[edge] (bottomcircle) -- node[below left]{$\frac{1}{3}$} (leftcircle);
\draw[edge] (rightcircle) -- node[below right]{$\frac{1}{2}$} (bottomcircle);
\draw[edge] (topcircle) -- node[above right]{$\frac{1}{2}$}(rightcircle);
\draw[edge] (bottomcircle) -- node[left]{$\frac{1}{3}$} (topcircle);
\draw[edge] (topcircle) to[loop above, looseness=15] node[above]{$\frac{1}{2}$} (topcircle);
\draw[edge] (leftcircle) to[loop left, looseness=15] node[left]{$\frac{1}{2}$} (leftcircle);
\draw[edge] (rightcircle) to[loop right, looseness=15] node[right]{$\frac{1}{2}$} (rightcircle);
\draw[edge] (bottomcircle) to[loop below, looseness=15] node[below]{$\frac{1}{3}$} (bottomcircle);
\end{tikzpicture}
}}
\subfloat[]{
\resizebox{0.49\columnwidth}{!}{
\begin{tikzpicture}[baseline]
% Nodes
\node[non-virtualnode] (topcircle) {$1$};
\node[non-virtualnode] (leftcircle) [below left = 1cm of topcircle] {$0$};
\node[non-virtualnode] (rightcircle) [below right = 1cm of topcircle] {$2$};
\node[non-virtualnode] (bottomcircle) [below right = 1cm of leftcircle] {$3$};
% Virtual Nodes
\node[virtualnode] (v_leftcircle1) [below left = 0.75cm of leftcircle] {\footnotesize{$0_1$}};
\node[virtualnode] (v_leftcircle2) [below = 0.75cm of v_leftcircle1] {\footnotesize{$0_2$}};
% --
% Lines
% --
\draw[edge] (leftcircle) -- node[above left]{$\frac{1}{2}$} (topcircle);
\draw[edge] (bottomcircle) -- node[below left]{$0$} (leftcircle);
\draw[edge] (rightcircle) -- node[below right]{$\frac{1}{2}$} (bottomcircle);
\draw[edge] (topcircle) -- node[above right]{$\frac{1}{2}$} (rightcircle);
\draw[edge] (bottomcircle) -- node[left]{$\frac{1}{3}$} (topcircle);

\draw[edge] (topcircle) to[loop above, looseness=15] node[above]{$\frac{1}{2}$} (topcircle);
\draw[edge] (leftcircle) to[loop left, looseness=15] node[left]{$\frac{1}{2}$} (leftcircle);
\draw[edge] (rightcircle) to[loop right, looseness=15] node[right]{$\frac{1}{2}$} (rightcircle);
\draw[edge] (bottomcircle) to[loop below, looseness=15] node[below]{$\frac{1}{3}$} (bottomcircle);

% Virtual Lines
% --
\draw[vedge] (bottomcircle) to[bend left=10] node[below left]{$0$} (v_leftcircle1);
\draw[vedge] (bottomcircle) to[bend left=15] node[below right]{$\frac{1}{3}$} (v_leftcircle2);
% |
\draw[vedge] (v_leftcircle1) to[bend left=5] node[left]{$1$} (leftcircle);
\draw[vedge] (v_leftcircle2) to[bend left=5] node[left]{$1$} (v_leftcircle1);
% --
\end{tikzpicture}
}}}
\caption{(a) Example of a delay-free $4$-worker network. Edges are labeled with column-stochastic weights $P_{i,j}$. (b) The same network augmented
with virtual nodes and edges (dashed blue lines). For readability, this figure only shows the virtual workers/edges used to model delays for messages transmitted to worker $0$; in the analysis, virtual nodes and edges are added for every worker. This particular example illustrates a message from $3$ to $0$ with a delay of $\taubar=2$.}
\label{fig:aug-graph}
\end{figure}

\textit{(i) Modelling delays:}
Recall that one can form a communication graph $\mathcal{G}(\mathcal{V}, \mathcal{E})$ with one vertex for each worker and with an edge set $\mathcal{E}$ containing a (directed) edge from $j$ to $i$ if $P_{i,j} \neq 0$.
In order to model message delays in analysis, the reference graph $\mathcal{G}(\mathcal{V}, \mathcal{E})$ is augmented with virtual nodes and edges that store information that has been transmitted but not yet received. Because the message delays are bounded, the number of virtual nodes and edges needed is finite.
Figure~\ref{fig:aug-graph} shows an example of such a graph augmentation for the delays along one edge.
Note that the message delays under this bounded delay model can still vary across edges, and can vary from one iteration to the next.
Graph augmentation techniques have also been used to study distributed averaging and agreement algorithms with communication delays~\cite{cao2008reaching,tsianos2011distributed,hadjicostis2014average,charalambous2015distributed}; additional work is needed to properly account for computation delays in asynchronous distributed optimization. 
In short, one can model asynchronous delay-prone message passing over the graph $\mathcal{G}(\mathcal{V}, \mathcal{E})$ as synchronous \emph{time-varying} message passing over the time-varying augmented graph $\tilde{\mathcal{G}}(\tilde{\mathcal{V}}, \tilde{\mathcal{E}}^{(k)})$.
One can also incorporate heterogeneous update rates into the analysis by multiplying the diagonal step-size matrix $\Gamma^{(k)}$ in equations~\eqref{eq:glob-synchronous-decentralized} and~\eqref{eq:glob-synchronous-col-decentralized} with a diagonal binary indicator matrix $\Delta^{(k)}$, with $i^\textup{th}$ diagonal $\delta^{(k)}_i$ indicating whether worker $i$ completed an update at global iteration $k$. We emphasize that graph augmentation techniques are only used for the purpose of analysis, to model decentralized systems with delays.

The augmented version of equation~\eqref{eq:glob-synchronous-decentralized} is
\begin{equation}
    \label{eq:glob-aug-synchronous-decentralized}
    \widetilde{W}^{(k + 1)} = \widetilde{P}^{(k)} \widetilde{W}^{(k)} - \widetilde{\Delta}^{(k)} \widetilde{\Gamma}^{(k)} \widetilde{S}^{(k)},
\end{equation}
where the augmented matrix $\widetilde{W}^{(k)}$ has dimensions $(\taubar + 1)n \times d$; \ie, one row containing the parameters at each worker and each virtual node at iteration $k$. We emphasize that this modeling is only used for analysis and is not needed to implement asynchronous decentralized methods. The rows of the $(\taubar+1)n \times d$ matrix $\widetilde{S}^{(k)}$ that correspond to virtual nodes are always equal to $0$.
Equation~\eqref{eq:glob-synchronous-col-decentralized} can be described similarly with respect to this enlarged state-space.

Note that the averaging matrix $\widetilde{P}^{(k)}$ in~\eqref{eq:glob-aug-synchronous-decentralized} is time-varying for two reasons: first, only those workers that are active at iteration $k$ perform an update, and second, message delays can vary across iterations.
Overall, this model reduces the time-varying and delay-prone dynamics of a decentralized asynchronous algorithm to the evolution of an augmented synchronous system.

\textit{(ii) Convergence to a consensus sequence:}
Note that~\eqref{eq:glob-aug-synchronous-decentralized} can be viewed as the
evolution of a perturbed Markov chain with state-transition matrix $\widetilde{P}^{(k)}$ and perturbations $\widetilde{\Delta}^{(k)} \widetilde{\Gamma}^{(k)} \widetilde{S}^{(k)}$.
Using standard tools from the Markov chain literature~\cite{wolfowitz1963products,seneta1981}, one can characterize the convergence rate of the iterates $w^{(k)}_i$ to a consensus sequence $\overline{w}^{(k)}$ using the joint spectral properties of the matrices $\widetilde{P}^{(k)}$.
The resulting bounds are often of the form
\begin{equation}
    \label{eq:perturbed-consensus}
    \norm{w^{(k)}_i - \overline{w}^{(k)}} \leq C \rho^k \norm{w^0_i} + C \sum^{k}_{\ell=0} \rho^{k}_i \norm{\delta^{(k - \ell)}_i \gamma^{(k - \ell)}_i s^{(k - \ell)}_i},
\end{equation}
for all $i\in[n]$, where $\rho\in (0,1)$ and $C\in (0,\infty)$ are constants that depend on the delays and graph connectivity, and $\{ \overline{w}^{(k)} \}$ is the consensus sequence.
It follows from~\eqref{eq:perturbed-consensus} that if the perturbations $(\delta^{(k)}_i \gamma^{(k)}_i s^{(k)}_i)$ tend to $0$, then all workers converge to the consensus sequence $\overline{w}^{(k)}$, even in the presence of arbitrary, uniformly bounded message and computation delays.
There are several different approaches in the literature for bounding $\rho$; see~\cite{assran2018asynchronous, zhang2019asyspa, tian2018asy} for details.
When the weight matrices $P^{(k)}$ are column-stochastic or doubly-stochastic, the consensus sequence $\overline{w}^{(k)}$ is typically defined as the network-wide average of the parameters at iteration $k$ (i.e., $\nicefrac{1}{n} \sum^n_{i=1}w^{(k)}_i$).

To prove that the iterates $w^{(k)}_i$ converge to the consensus sequence using~\eqref{eq:perturbed-consensus}, one must show that the perturbations (gradient-based updates) $(\delta^{(k)}_i \gamma^{(k)}_i s^{(k)}_i)$ tend to $0$.
When using a diminishing step-size (\ie, $\gamma^{(k)}_i \to 0$), this is a trivial result (as long as the search directions remain bounded).
When using a constant step-size, the conditions for consensus, namely $(\delta^{(k)}_i \gamma^{(k)}_i s^{(k)}_i)$ converging to $0$ for all $i\in [n]$, and the conditions for optimality, namely $\overline{w}^{(k)}$ converging to a minimizer, are often tightly interdependent.
Gradient-tracking methods using a constant step-size, such as \textsc{Asy-Sonata}, have this interdependence, and typically show consensus and optimality simultaneously using~\eqref{eq:perturbed-consensus} and the \emph{small-gain theorem}~\cite{nedic2017achieving}.
In brief, the \emph{small-gain-theorem} says that if, for all positive integers $K$, there exists $\lambda \in (0,1)$, finite constants $C_1, C_2 \geq 0$, and gains $G_1, G_2 \geq 0$ with $G_1 G_2 < 1$, such that
\[
    \sup_{k \leq K} \frac{\norm{s^{(k)}_i}}{\lambda^k} \leq \sup_{k \leq K} G_1 \frac{\norm{w^{(k)}_i - \overline{w}^{(k)}}}{\lambda^k} + C_1,
\]
and
\[
    \sup_{k \leq K} \frac{\norm{w^{(k)}_i - \overline{w}^{(k)}}}{\lambda^k} \leq \sup_{k \leq K} G_2  \frac{\norm{s^{(k)}_i}}{\lambda^k} + C_2,
\]
then both $\norm{w^{(k)}_i - \overline{w}^{(k)}}$ and $\norm{s^{(k)}_i}$ converge to $0$ at a linear rate characterized by the sequence $\{\lambda^k\}$.
It is relatively straightforward to generalize the small-gain-theorem to characterize other convergence rates as well; e.g., proving sublinear convergence by replacing the sequence $\{ \lambda^k \}$ in the denominators with a sublinearly convergent sequence $\{r_k\}$.

\textit{(iii) Convergence of consensus sequence to a minimizer:}
We will now describe a general approach for proving convergence of the consensus sequence in a way that provides some intuition into the convergence behaviour of asynchronous decentralized optimization methods.

Due to the presence of the binary indicator $\delta^{(k)}_i$ in the gradient-based updates in~\eqref{eq:glob-aug-synchronous-decentralized}, one cannot guarantee a contraction with respect to the global objective at sufficiently large iterations $k$.
Intuitively, some workers may take gradient steps that move the parameters away from the global minimizer.
The key observation is that, while each iteration may not produce a descent direction, the sum of the gradient-based updates $\sum_k \sum^n_{i=1} \delta^{(k)}_i \gamma^{(k)}_i s^{(k)}_i$ over sufficiently many consecutive iterations may point in a descent direction when the computation delays are bounded. For example, for \agp it can be shown that this cumulative gradient vector points in a descent direction with respect to the re-weighted minimizer defined in~\eqref{eq:rw-prob}; see~\cite[Lemmas~2 and 3]{assran2018asynchronous}.

Typically, after obtaining a contraction result over a finite-time horizon, standard tools from the optimization literature can be applied to obtain convergence of the consensus sequence.

\textit{Validity of the bounded delay assumption.} All analysis techniques presented in this article assume arbitrary, uniformly bounded (but possibly time-varying) message and computation delays.
In practice, we can control the upper bound on the delays by using tools described in Section~\ref{sec:architectures}.
If a worker has not received a message from its neighbours in over $\tau$ iterations, it blocks and waits to receive a message.

\subsection{Non-convex objectives}
While the discussion in this section has largely focused on convex objectives, much of the analysis techniques can be extended to non-convex objectives with minor alterations.
Specifically, steps (i) and (ii) in the analysis (modelling delays and proving convergence of the optimization iterates to a consensus sequence), remain unchanged. Step (iii), proving convergence of the consensus sequence to a minimizer, is the only part that requires adjustment to account for non-convex objectives.
In step (iii), one must obtain the contraction result over a finite-time horizon without making use of the (sub)gradient inequality; instead, it is common to make use of a Taylor-series expansion to express the relationship between the change in the objective error after taking an optimization step, and the expected descent provided by the stochastic search-direction over a finite-time horizon.

As an example, the analysis of the \osgp method for smooth non-convex objectives with stochastic gradients in~\cite{assran2019stochastic} uses this general proof sketch.
Suppose the message delays are uniformly bounded (\ie, there exists a $\taubar > 0$ such that $\tau^{(k)}_{i,j} \leq \taubar$ for all $k$ and $i,j \in [n]$).
If \osgp is run for $K$ iterations and all workers use a constant step-size $\gamma \coloneqq \sqrt{\nicefrac{n}{K}}$, then each worker is guaranteed to converge to a stationary point of~\eqref{eq:erm} when the objectives $f_i$ are non-convex and $L$-smooth~\cite{assran2019stochastic}
\begin{equation}
\label{eq:osgp-nonconvex}
    \frac{1}{n} \frac{1}{K} \sum^K_{k=0} \sum^n_{i=1} \norm{\nabla f_i(z^{(k)}_i)}^2 \leq \mathcal{O}\left(\frac{1}{\sqrt{nK}}\right).
\end{equation}
Remarkably, \osgp converges to a stationary point of smooth non-convex functions with the same order-wise iteration complexity as centralized \sgd.
Since the analysis for \osgp handles general (strongly-connected) digraphs, and arbitrary (but bounded) time-varying message delays, equation~\eqref{eq:osgp-nonconvex} also provides a bound on the convergence rate for \sgp (synchronous delay-free setting), and \allreduce \sgd (synchronous delay-free all-to-all setting).

\subsection{Example: Training a deep neural network}

Next we provide an illustrative example of how some of the algorithms mentioned in Section~\ref{sub-sec:decentralized-methods} can be used to speed up training of a deep convolutional neural network model on an image classification task.
Each worker is a server consisting of $40$~CPU cores and $8$~GPUs. The servers are interconnected via a $10$ Gbps Ethernet network. We train a ResNet-50~\cite{he2016deep} model containing roughly $25$ million optimizable parameters to classify images in the ImageNet dataset~\cite{russakovsky2015imagenet}, made up of over $1$ million images and $1000$ different image classes. 

We compare \osgp, \sgp, and \allreduce \sgd. Recall that \allreduce \sgd is a synchronous method that exactly synchronizes all workers after every update, mathematically equivalent to taking $P_{i,j}^{(k)} = 1/n$ for all $i,j \in [n]$. \sgp is a synchronous decentralized method, and \osgp is an asynchronous decentralized method whose analysis allows for delayed messages.
Both \osgp and \sgp use a time-varying communication graph sequence $P_{i,j}^{(k)}$ with $1$ out-neighbour per node; \ie, after each update, a worker transmits a message to just one other worker.
See~\cite{assran2019stochastic} for additional details about the experimental setup.

\begin{figure*}[t]
\centering
\subfloat[.3\textwidth][Training loss vs. iterations, using $32$ workers ($256$ GPUs and $1280$ CPUs)]{\includegraphics[width=.3\textwidth]{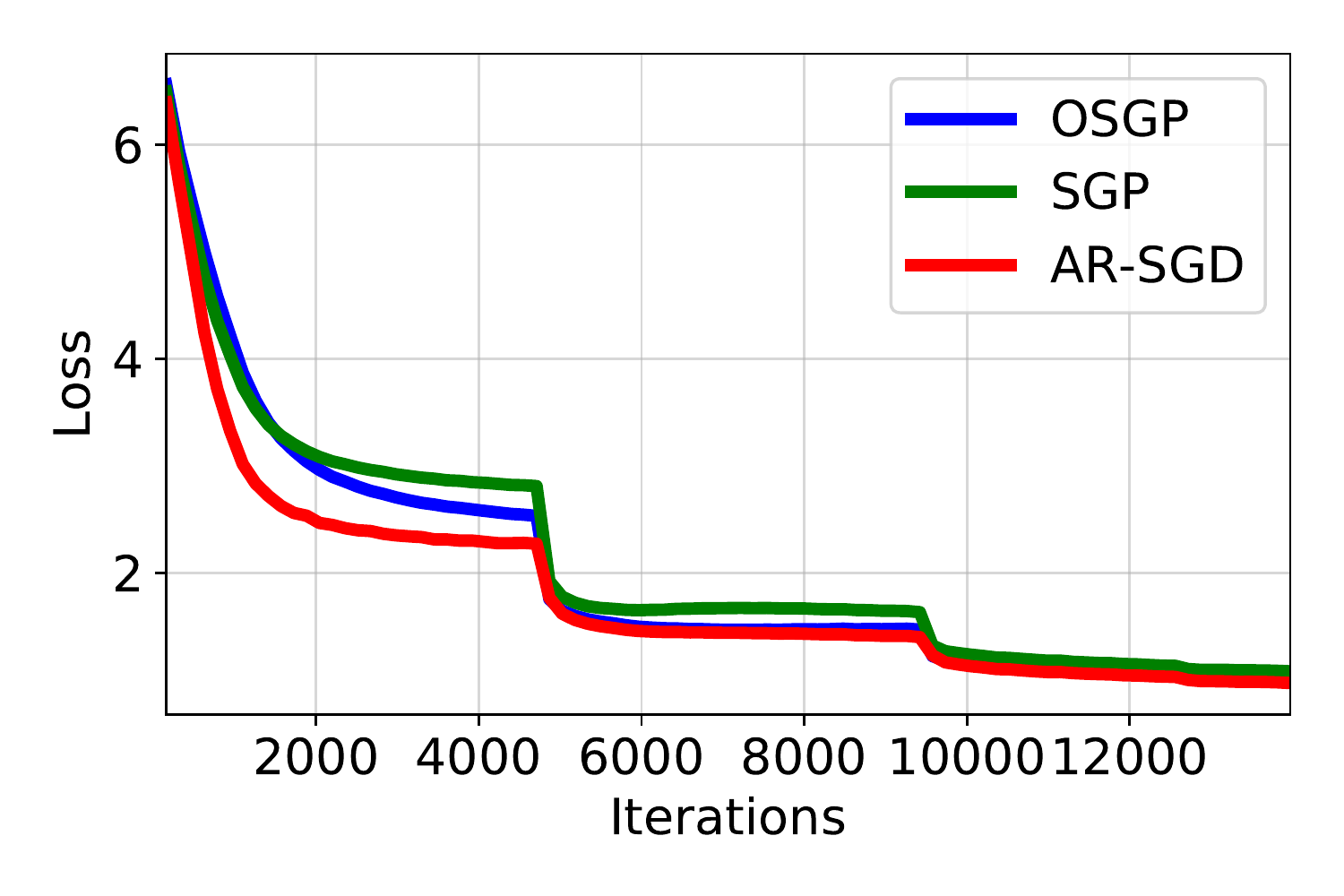}} \quad \quad
\subfloat[.3\textwidth][Training loss vs. time, using $32$ workers ($256$ GPUs and $1280$ CPUs)]{\includegraphics[width=.3\textwidth]{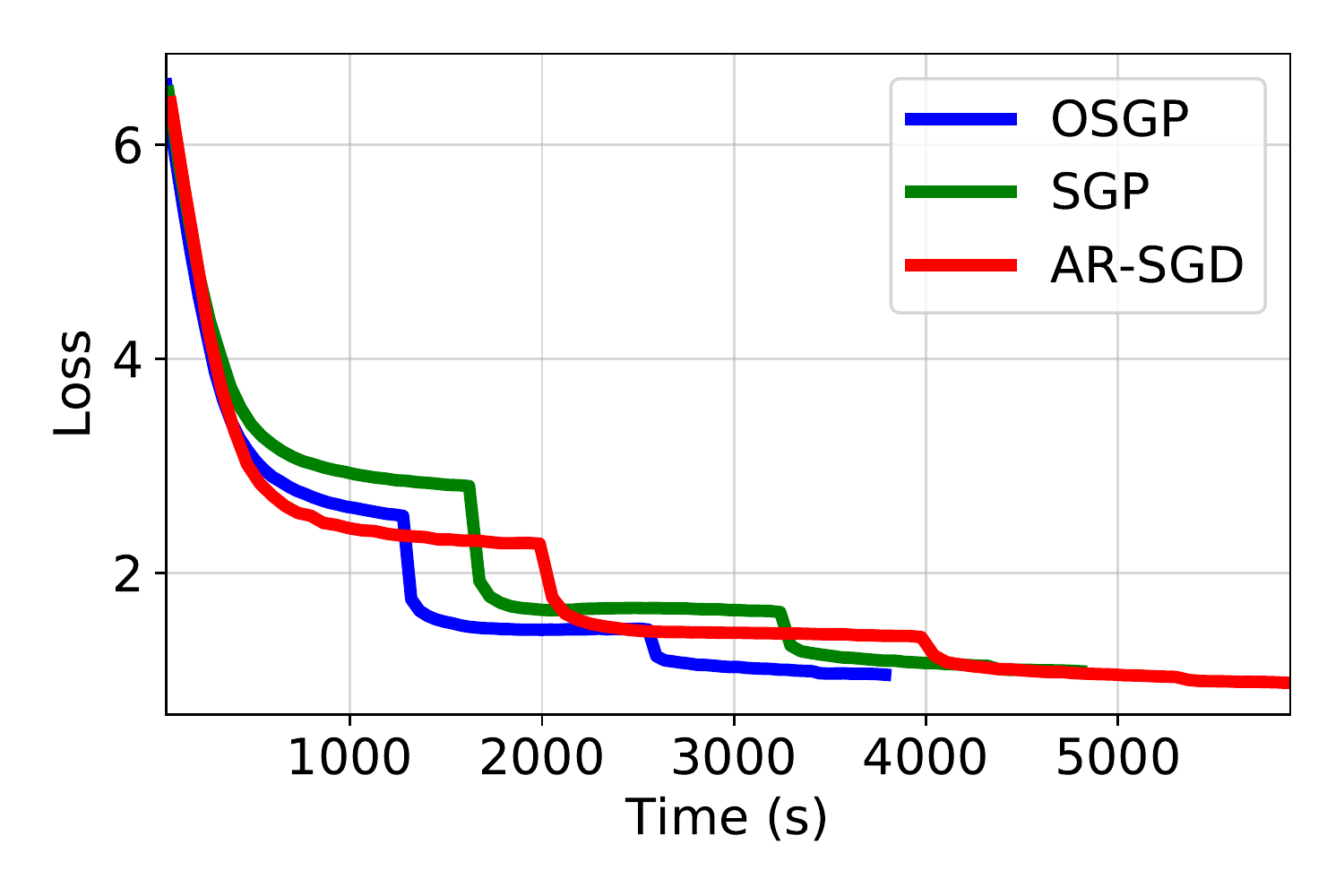}} \quad \quad
\subfloat[.3\textwidth][Time-per-iteration (ms/itr) vs. number of workers]{\includegraphics[width=.3\textwidth]{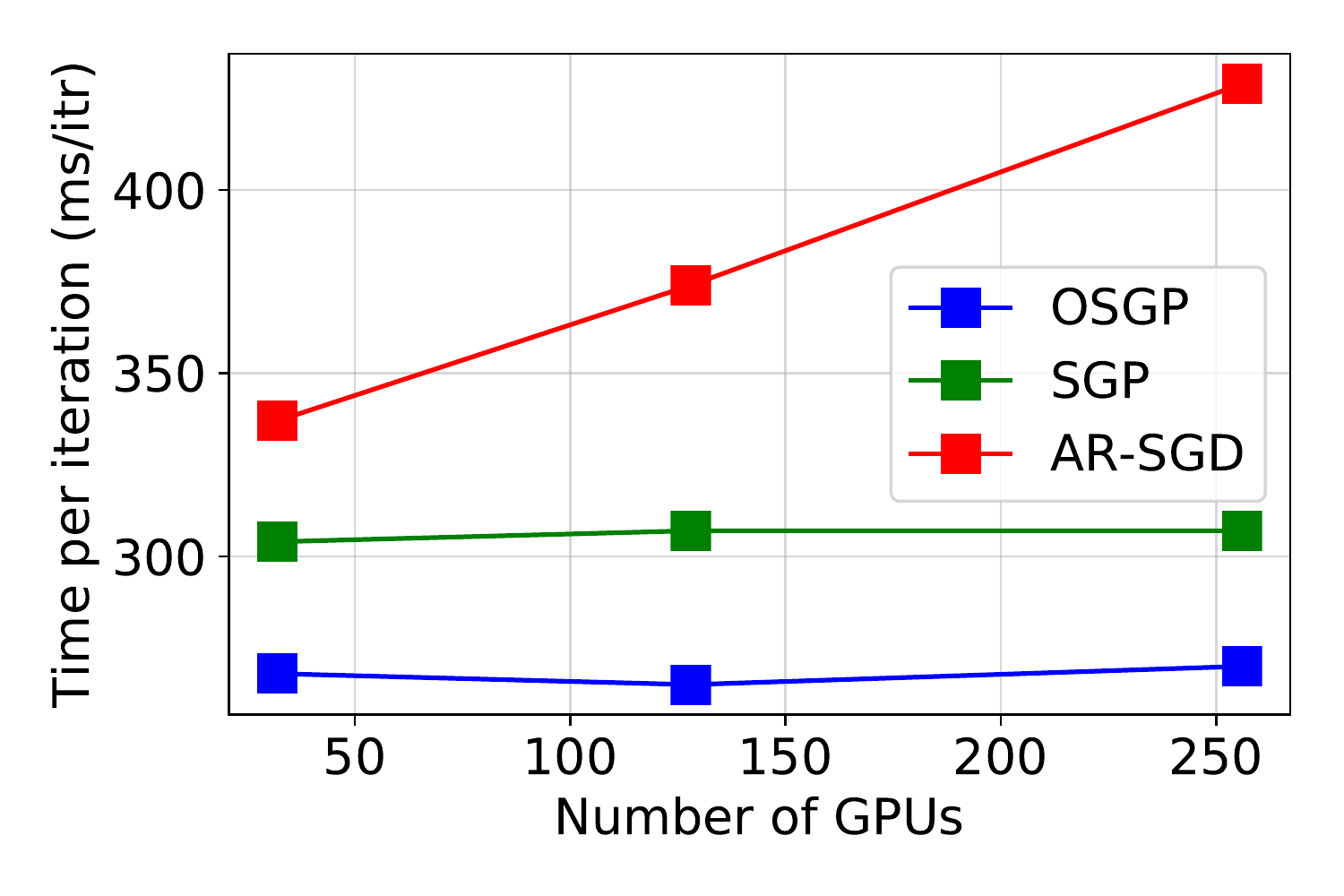}}
\caption{Training a ResNet-50 convolutional neural network on the ImageNet classification task over a network of servers. Each worker is an entire server consisting of $40$ CPUs and $8$ GPUs. Workers are interconnected by a $10$ Gbps ethernet network. (a)/(b) The \emph{average} training loss versus the number of iterations/wall-clock time when training with 32 workers. Shaded around each line is the max.~and min.~error achieved by any given worker in that iteration. (c) Observing how each worker's time per iteration scales as we increase the number of workers (proportional to the number of GPUs).}
\label{fig:imagenet}
\end{figure*}

Figure~\ref{fig:imagenet} (a) shows the loss, $f(w)$, as a function the number of iterations when training with $32$ workers ($256$ GPUs and $1280$ CPU cores). 
Figure~\ref{fig:imagenet} (b) shows the same loss as a function of the wall-clock time.
Although \osgp and \sgp converge at slower iteration-wise rates than \allreduce \sgd, Figure~\ref{fig:imagenet} (b) illustrates that \osgp has significantly reduced training time by mitigating the synchronization and communication overhead.
Figure~\ref{fig:imagenet} (c) shows the time per iteration as a function of the number of workers in the system, for $n=4$, $16$, and $32$ (\ie, $32$, $128$, and $256$ GPUs).
The time-per-iteration for both \osgp and \sgp remains relatively constant since the communication overhead is always fixed.
The time-per-iteration of \allreduce \sgd increases since the synchronization and communication costs increase with the number of workers. In short, the asynchronous decentralized method \osgp optimizes $f(w)$ faster than the synchronous methods (in terms of wall-clock time), and exhibits better scaling.

\section{Conclusions and directions}\label{sec:conc}

While asynchronous decentralized optimization algorithms have shown promising
results on large-scale machine learning benchmarks~\cite{lian2017can,
lian2018asynchronous, assran2019stochastic, wang2019slowmo}, there is still much
work to be done in understanding the convergence properties of these methods.
Section~\ref{sub-sec:decentralized-methods}
outlines some of the known convergence results, but the constants in these rates
are typically large and do not accurately reflect the scaling behaviour of the
algorithms.  For instance, it is not clear how well these constants reflect the
true dependency of the convergence rates on the number of workers, the
communication graph, or the delays due to asynchrony. Some recent works such
as~\cite{nedic2018network, olshevsky2019asymptotic} try to provide a closer look
at the constants in the convergence rates of synchronous decentralized
optimization algorithms.

Another challenge involves incorporating non-linear gradient-based updates.
Optimization methods using non-linear momentum-based updates are commonly used in deep learning~\cite{kingma2014adam,hinton2012neural}.
If the gradient-based update is non-linear, then it may not be possible to guarantee that the consensus sequence converges to a minimizer of~\eqref{eq:erm}.
This issue applies to both synchronous and asynchronous decentralized optimization algorithms.
In practice, one typically observes drastic degradation in performance, relative
to centralized methods, when na\"{i}vely applying decentralized optimization algorithms with non-linear gradient-based updates to large-scale deep learning tasks.
One recent work~\cite{wang2019slowmo} tackles this issue in the synchronous case by periodically incorporating global synchronization between agents.
Incorporating non-linear gradient-based updates into both synchronous and asynchronous multi-agent optimization algorithms is still an open problem.

Our discussion focused on analysis under a partially asynchronous delay model, where information delays are only assumed to be bounded, and thus did not cover other work which assumes that delays follow more specific probabilistic models~\cite{cannelli2016asynchronous,mansoori2017superlinearly}. Such assumptions may lead to tighter bounds when they accurately reflect the salient properties of the underlying system, but verifying the validity of such models is more challenging in practice. In contrast, bounded information delays can be enforced algorithmically, albeit potentially at the cost of some idling.

Lastly, although the inconsistent read perspective is a convenient abstraction for analysis of parallel optimization algorithms in shared memory, it is not an accurate description of the behavior of current multi-core systems with non-uniform memory access. In these systems, frequent concurrent operations on the same elements in shared memory create contention and reduce the efficiency of cache hierarchies. Instead, emerging high-performance algorithms for multi-processors, such as~\cite{IMP:19,ZHA:16} bear striking resemblance with the decentralized methods described in this paper: cores operate on local (inconsistent) copies of the decision vector and coordinate to guarantee global convergence.  We believe that there is a significant scope for designing new algorithms tailored to the specifics of NUMA architectures instead of adapting algorithms designed with simpler hardware abstractions.

\bibliographystyle{IEEEtran}
\bibliography{references.bib}

\end{document}